\documentclass[manuscript,nonacm]{acmart}

\setcopyright{acmlicensed}
\copyrightyear{2026}
\acmYear{2026}
\acmDOI{XXXXXXX.XXXXXXX}

\author{Snehalkumar `Neil' S. Gaikwad}
\email{gaikwad@cs.unc.edu}
\correspondingauthor
\affiliation{%
  \institution{Society-Centered AI Lab, Department of Computer Science, School of Data and Information Sciences, University of North Carolina at Chapel Hill}
  \city{Chapel Hill}
  \state{North Carolina}
  \country{USA}
}

\author{Punya Aragula}
\affiliation{%
  \institution{Society-Centered AI Lab, School of Data and Information Sciences, University of North Carolina at Chapel Hill}
  \city{Chapel Hill}
  \state{North Carolina}
  \country{USA}
}

\author{Shiva Kruthi Palpanoori}
\affiliation{%
  \institution{Society-Centered AI Lab, School of Data and Information Sciences, University of North Carolina at Chapel Hill}
  \city{Chapel Hill}
  \state{North Carolina}
  \country{USA}
}

\author{Pratik Patekar}
\affiliation{%
  \institution{Society-Centered AI Lab, School of Data and Information Sciences, University of North Carolina at Chapel Hill}
  \city{Chapel Hill}
  \state{North Carolina}
  \country{USA}
}

\renewcommand{\shortauthors}{Gaikwad et al.}

\usepackage{booktabs}
\usepackage{placeins}
\usepackage{caption}
\usepackage{fontawesome5}
\usepackage[utf8]{inputenc}
\usepackage{float}
\usepackage{multirow}
\usepackage{longtable}
\AtBeginDocument{%
  }
\usepackage{kvoptions}
\usepackage{fancybox}
\usepackage{setspace}
\usepackage{longtable}
\usepackage{booktabs}
\usepackage{array}
\usepackage{ragged2e}
\usepackage{tabularx}
\usepackage{booktabs}
\usepackage{placeins}
\usepackage[most]{tcolorbox}
\usepackage{adjustbox}
\usepackage{ragged2e}

\definecolor{blockgray}{gray}{0.95}
\definecolor{ftblue}{RGB}{235, 241, 246}
\definecolor{ftaccent}{RGB}{40, 75, 110}
 
\newtcolorbox{promptbox}[2][]{
    colback=ftblue,
    colframe=ftaccent,
    rounded corners,
    arc=2mm,
    boxrule=0.4pt,
    borderline west={2pt}{0pt}{blue!70!black},
    fonttitle=\bfseries,
    title={{#2}},
    before upper=\ttfamily\normalsize,
    breakable,
    width=\linewidth,
    #1
}

\newcolumntype{Y}{>{\RaggedRight\arraybackslash}X}
\newcolumntype{L}[1]{>{\RaggedRight\arraybackslash}p{#1}}
\makeatletter
\SetupKeyvalOptions{family=PB, prefix=PB@}
\DeclareStringOption[promptbody]{placeholder}
\DeclareStringOption[prompttitle]{title}
\DeclareStringOption[blue!50!black]{bordercolor}
\DeclareStringOption[blue!2!white]{fillcolor}
\DeclareStringOption[blue!70!black]{titlecolor}
\DeclareStringOption[1.5pt]{linewidth}
\DeclareBoolOption[true]{showtitle}
\ProcessKeyvalOptions{PB}

\makeatother

\begin{document}

\title{Between Niti and Nyaya: A Society-Centered Framework for AI Governance}
 
\begin{abstract}
Governments are making decisions about AI before they can measure its public effects, and those decisions determine whether a person can learn that AI influenced an outcome, challenge it, and see it corrected. Research has documented what AI governance produces: principles, strategies, registers, and audits. However, we lack shared language for understanding AI policy while it is still being made. We introduce the Society-Centered AI Governance Framework, which gives policymakers, researchers, and the public a way to assess instruments in any jurisdiction by whether a person can get notice, challenge a decision, and see it corrected. It reads a jurisdiction's instruments together by date and binding status, separates an agency duty from a right a person can invoke, and traces which later instrument must enforce each commitment. We evaluated it in North Carolina using interviews with nine AI Leadership Council members, conducted before and after the state's 2026 AI roadmap, and a provision-level reading, with human-reviewed model-assisted coding, of 49 AI governance documents from 19 states. For the state's own AI systems, we find visibility-first governance: visibility duties are in effect, notice exists as an agency duty with no right a person can invoke, and AI-specific challenge and correction remain advisory. The roadmap assigns owners and dates to the missing protections, so the gaps are documented but not yet closed. Recent federal preemption proposals focus on AI development, so recourse for a state's own systems is likely to remain a state decision. We identify "legitimacy before protection" as a testable risk of visibility-first governance and name the milestones that would test it. More broadly, judging AI policy by who can act on what it makes visible lets the field assess recourse while policy is still forming, before harm occurs.
\end{abstract}



\maketitle

\section{Introduction}
\label{sec:intro}

In December 2020, the U.S. federal government required agencies to inventory their AI use cases~\cite{eo13960}. A 2024 federal memorandum added notice, opt-out, human review, and appeal~\cite{omb2024m2410, cdt2025omb}, and its April 2025 replacement kept inventories, appeal, and human oversight and removed notice and opt-out~\cite{omb2025m2521, cdt2025omb, wiley2025omb}. Across those changes, the record of what the government runs lasted, while the requirement to tell a person that AI was used came and went. How does AI policy take shape, and what does it deliver for the people it affects? Governments are making decisions about AI in benefits \cite{eubanks2018automating}, child welfare \cite{saxena2021framework}, and public safety \cite{richardson2019dirty} before they can measure its public effects~\cite{lawrence2023bureaucratic, feng2025sociotechnical, engstrom2020government}. Policymakers face what the International AI Safety Report calls an evidence dilemma: acting too early can entrench ineffective interventions, and waiting for conclusive data can leave society vulnerable~\cite{bengio2026international}. Those decisions pass through governors' offices, advisory councils, legislatures, and agencies, and a published plan rarely shows whose concerns shaped it or what agencies could staff. A benefits claimant or a family in a child welfare case meets the result as a notice or an appeal deadline~\cite{citron2008technological, eubanks2018automating, karusala2024understanding}, and these decisions determine whether that person can learn that AI influenced an outcome, challenge it, and see it corrected~\cite{lyons2022s, karusala2024understanding}. These questions now sit inside an active policy debate: U.S. state AI laws rose from one in 2016 to 131 in 2024~\cite{stanford2025aiindex}, Congress has not enacted preemption of state AI laws~\cite{ropes2026limits, casrai2026preemption}, and Colorado replaced its AI Act in May 2026~\cite{eckert2026colorado, colorado_sb26_189}. Much of this debate concerns companies, but states also deploy AI. The preemption proposals we reviewed focus on development and exempt state government procurement~\cite{ropes2026limits, casrai2026preemption}, so what a resident can do about a state's own AI is likely to remain a state decision.

Prior work has examined ethics principles~\cite{jobin2019global, morley2020initial}, national and state strategies~\cite{bareis2022talking, migineishvili2026regulating}, and registers and audits~\cite{das2026bureaucratic, groves2024auditing}. Planning and policy theory has long studied this kind of decision-making, from Lindblom's muddling through~\cite{lindblom1959science} and Simon's bounded rationality~\cite{simon1955behavioral} to Altshuler on limited authority~\cite{altshuler1965goals}, Davidoff on plural public interests~\cite{davidoff1965advocacy}, and Sanyal on institutional resistance~\cite{sanyal2005planning}. To date, AI governance research has largely examined what governance produces, and it has drawn less on planning theory's account of how commitments become binding. However, the core challenge is that we lack shared language for AI policy while it is still being made. Hard law and soft law~\cite{alanoca2025comparing} tell us whether an instrument binds, and they leave open what it lets a person do and which later instrument must carry it out. This matters because officials decide before anyone can observe the lives Sen asks us to judge~\cite{sen2009idea}, diffuse harms rarely produce the organized parties and named agents that Stone's account centers on~\cite{stone2012policy}, and one instrument can inform the state while giving an affected person no way to respond.

\begin{figure*}[t]
\centering
\includegraphics[width=\textwidth]{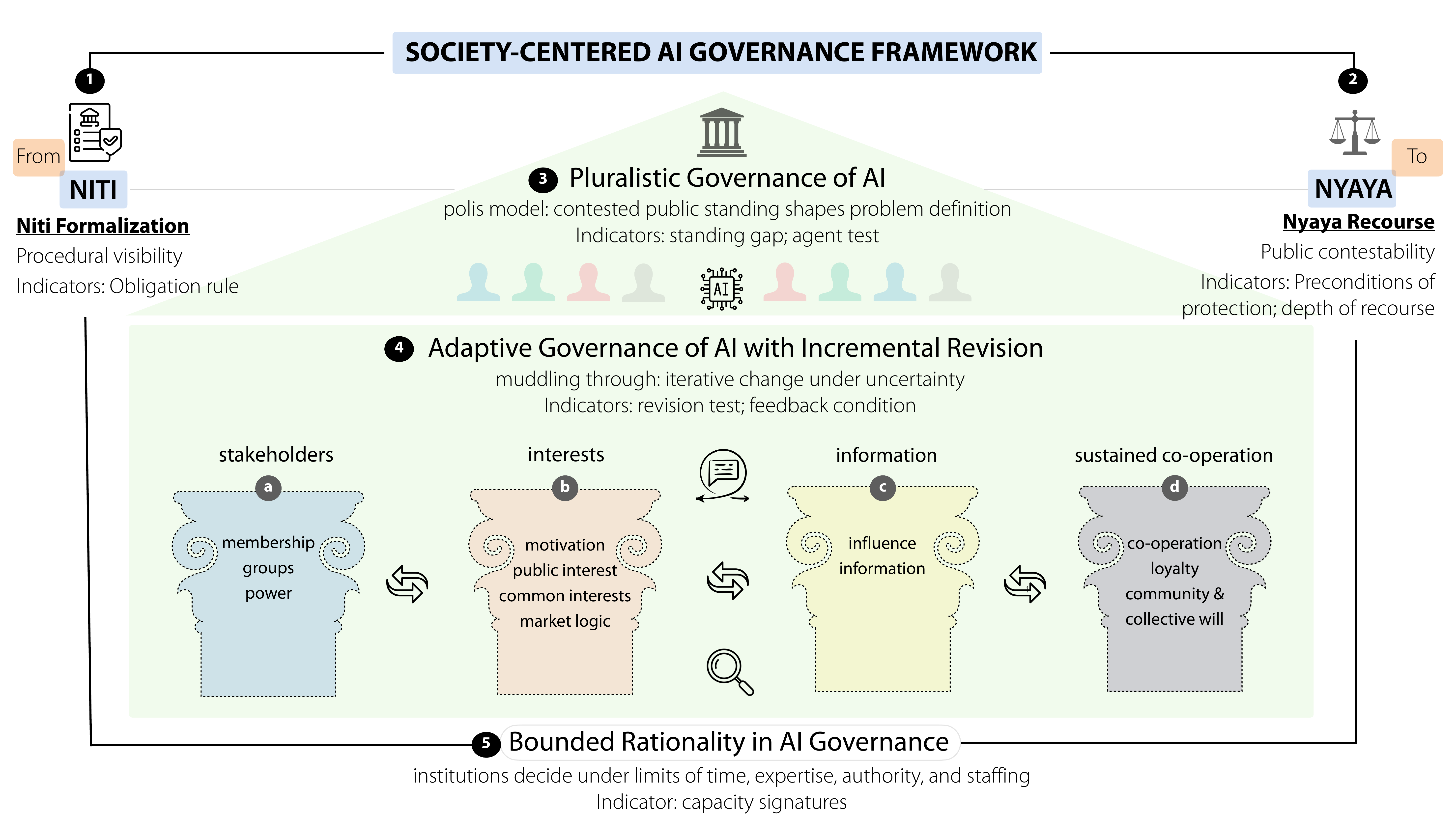}
\caption{The Society-Centered AI Governance Framework. Niti Formalization (1) records what institutions build, and Nyaya Recourse (2) records what reaches affected people. Pluralistic Governance of AI (3), Adaptive Governance of AI with Incremental Revision (4), and Bounded Rationality in AI Governance (5) explain how policy formation settles the relation between the two. The four pillars (a to d) group the subcodes of component 3.}\label{fig:framework}
\end{figure*}

In this paper, we draw on planning and policy theory to introduce the \textit{Society-Centered AI Governance Framework}, a lens for understanding how AI policy forms, how officials make decisions within it, and what those decisions deliver. The framework follows a jurisdiction's full set of instruments, judges them by what an affected person can do, explains how competing interests, incremental revision, and limited capacity shape the relation between the two, and offers four questions for testing any governance instrument. It serves three audiences: policymakers can test an instrument before adoption, researchers can compare jurisdictions before outcome data exists, and the public can ask what a plan lets a person do. The framework is general: its unit of analysis is the set of instruments that govern a class of AI systems, so it applies to a nation, a state, a city, or an international body. Three concepts anchor it: \textit{procedural visibility} and \textit{public contestability} separate what an instrument lets the state see from what it lets a person challenge, the \textit{binding sequence} records whether each instrument took effect and whether it creates an agency duty or a right a person can invoke, and \textit{handoff specification} asks which later instrument will enforce a commitment.

We evaluate the framework in North Carolina, which acted early and in the open: AI governance instruments appear across executive, legislative, and judicial contexts, and every Roadmap recommendation has an owner and a date. Its detailed public record lets us test the framework provision by provision, and we also read federal, European, and Canadian cases through it (Section~\ref{sec:beyond}). We interviewed nine members of the state's AI Leadership Council, five before and four after the July 2026 release of its \textit{Statewide AI Strategic Roadmap}, and we read their reasoning against an AI governance corpus of 49 documents from 19 states, coded with model assistance and human review.

Our general claim is that a government can make AI visible to itself well before it gives affected people anything to invoke, and that reading instruments one at a time hides the distance between the two. In North Carolina, we observe a pattern we call \textit{visibility-first governance} for the state's own AI systems: duties that make those systems visible to the state are in effect, and AI-specific protections for challenge and correction are at earlier stages, whether advisory, internal to agencies, or still to be written (Table~\ref{tab:matrix}). The state's Responsible Use Framework shows the distinction: it directs agencies to notify people who may be impacted by AI and says that it creates no enforceable right (NC, ID-6). Each branch acted within its authority, and the Roadmap assigns owners and dates to its next steps, so the record shows a documented, incomplete sequence, which divided authority and limited capacity help explain. Across harms, enacted and proposed enforcement provisions address offenders, creators, and developers, and the Roadmap addresses harms from the state's own systems through recommendations scheduled for 2026 and 2027. We identify \textit{legitimacy before protection} as a testable risk: inventories and a public roadmap may build trust before a binding challenge procedure exists, and we state the milestones that could test it. These findings matter now because any government can decide who may invoke the duties it has already written.

\vspace{3em}
We make four contributions. 
\begin{itemize}
  \item \textbf{A shared language for AI policy in formation.} The Society-Centered AI Governance Framework names what existing terms leave open: procedural visibility and public contestability, agency duties and affected-person rights, the binding sequence, and handoff. Policymakers, researchers, and the public can apply it in any jurisdiction before outcome evidence exists (Section~\ref{sec:vocabulary}).
  \item \textbf{An account of visibility-first governance.} A provision-level reading of North Carolina's instruments in a 49-document corpus, read with nine Council interviews, shows visibility duties in effect while AI-specific challenge and correction for the state's own systems stay advisory. A notice duty already in effect gives a person nothing to invoke (Section~\ref{sec:findings}).
  \item \textbf{A temporal, mixed method for studying plans.} Interviews around a plan's release, paired with model-assisted coding and a provision-level reading, show what documents alone do not: bounded rationality surfaced in the interviews, and the coding traced it across the corpus (Section~\ref{sec:method}).
  \item \textbf{A testable mechanism.} Legitimacy before protection predicts how trust relates to recourse, and dated milestones separate it from planned sequencing and deferral (Section~\ref{sec:legitimacy}).
\end{itemize}

\section{Related Work}
We draw on three traditions: AI governance and policy implementation, algorithmic accountability and participation, and planning and policy theory.
 
\subsection{AI Governance and Policy Implementation}
 
AI governance research first documented principles, and a review of 84 ethics guidelines found convergence around transparency, fairness, responsibility, privacy, and non-maleficence~\cite{jobin2019global}. Attention then moved to tools~\cite{kuehnert2025and} and to comparisons of how governments allocate authority~\cite{bareis2022talking, bogiatzis2024beyond, schiff2023looking}. AI use in health care, labor, and education crosses statutes and oversight offices~\cite{bogiatzis2024beyond, cihon2020fragmentation}, one term can carry voluntary guidance or a binding duty~\cite{bateyko2025one}, and competing descriptions frame AI as a productivity tool, public-service infrastructure, and a security risk~\cite{bakiner2023pluralistic, ulnicane2023politics}. Work on generative AI argues that governance must adjust as systems change~\cite{reuel2024generative, hacker2023regulating}, and HCI, which has long studied technology policy~\cite{lazar2015public, jackson2014policy, spaa2019understanding}, now has an agenda for sociotechnical AI governance~\cite{feng2025sociotechnical}. A review of public records verified implementation of fewer than 40 percent of 45 federal AI requirements~\cite{lawrence2023bureaucratic}, and nine of 407 AI-related federal funding opportunities contained explicit AI guardrails~\cite{bateyko2025one}. To date, these studies have largely examined one kind of output or one stage of implementation. However, a benefits claimant's recourse may depend at once on decisions made in a governor's office, an advisory council, a legislative committee, and an agency budget. We therefore study these decisions as one process, following each commitment from the people who made it to whether it took effect.
 
\subsection{Algorithmic Accountability, Contestability, and Participation}
 
Accountability research treats audit and recourse as arrangements built from public records~\cite{diakopoulos2016accountability}, and contestability research follows the affected person through explanation, review, and appeal~\cite{vaccaro2019contestability, lyons2022s, karusala2024understanding}. Audit frameworks describe how organizations examine systems against stated commitments~\cite{raji2020closing, metcalf2021algorithmic}, and institutional pressures shape what impact assessments produce~\cite{selbst2021institutional}. Legal scholarship asks what due process requires when government decisions are automated~\cite{citron2008technological}. Most participatory AI efforts sit at the informing and consulting rungs of Arnstein's ladder~\cite{corbett2023power, arnstein1969ladder}, and the benefits of participation depend on context~\cite{birhane2022power}. New York City's bias audit regime established procedural compliance and left remediation to other mechanisms~\cite{groves2024auditing}, and a study of Canada's federal AI register describes it as offering visibility without contestability~\cite{das2026bureaucratic}. Both show what one instrument offers the public, and to date contestability research has largely studied a system or a single instrument. We extend it to the full set of instruments that govern a system and ask who can invoke each one.
 
\subsection{Policy Process, Planning Theory, and Institutional Legitimacy}
 
Public policy research describes a linear model from agenda to implementation~\cite{kingdon1995agendas, sabatier1986top, hupe2014happens}, and planning theory adds a longer record. Altshuler showed that comprehensive planners work without full authority or knowledge~\cite{altshuler1965goals}, Davidoff argued that no single public interest exists~\cite{davidoff1965advocacy}, Innes showed that consensus depends on who takes part~\cite{innes1996planning}, and Sanyal describes planning as the anticipation of institutional resistance~\cite{sanyal2005planning}. India's Five-Year Plans show the distance that can open between a plan's design and its outcomes~\cite{menon2022planning}, and AI adds the features of Rittel and Webber's wicked problems~\cite{rittel1973dilemmas}, since definitions vary across regulatory contexts~\cite{alanoca2025comparing, depaula2024regulating}. Meyer and Rowan argue that organizations adopt structures that signal conformity to accepted rules~\cite{meyer1977institutionalized}, Edelman traces how organizations give ambiguous civil rights law its working meaning~\cite{edelman1992legal}, and Yiftachel examines how plans can legitimate state action~\cite{yiftachel1998planning}. Organizational theory explains how institutions act under uncertainty: they search near existing solutions, reuse routines~\cite{cyert1963behavioral, nelson1985evolutionary}, and address what attention allows~\cite{ocasio1997towards}. Implementation research often follows a statute through the agencies that administer it, and AI rules are split across instruments with different legal force. Existing AI governance frameworks address organizational risk management~\cite{nist2023airmf}, regulation across jurisdictions~\cite{alanoca2025comparing}, and contestability within AI systems~\cite{alfrink2023contestable}. Our Society-Centered AI Governance Framework examines policy formation by tracing how commitments become binding and what affected people can do under them. It draws on Lindblom's muddling through~\cite{lindblom1959science}, Stone's polis~\cite{stone2012policy}, and Simon's bounded rationality~\cite{simon1955behavioral} to explain how institutions make these decisions. Following Sen~\cite{sen2009idea}, it provides pathways to evaluate the resulting instruments from the position of affected people.

\section{The Society-Centered AI Governance Framework}
\label{sec:vocabulary}

We call the framework \textit{``society-centered''} because it evaluates governance instruments by what they let affected people do: identify AI's public effects, challenge them, and seek correction. It tests any governance instrument through four public questions: Who is affected? What harm is tracked? Where can a person complain? Who has authority to correct it? Table \ref{tab:fourquestions} applies these questions to six North Carolina instruments. The first two concern what the state records, and an instrument that answers only them has procedural visibility. The last two concern what an affected person can do, and each answer adds public contestability.

Five components organize this work (Figure~\ref{fig:framework}; Table~\ref{tab:framework}). Niti Formalization and Nyaya Recourse draw on Sen's distinction between \textit{Niti}, formal rules and institutions, and \textit{Nyaya}, what those institutions deliver to people~\cite{sen2009idea, sen2008idea}. Outcomes cannot yet be observed while policy forms, so we hold \textit{Nyaya} as the standard and operationalize it as the availability of mechanisms: notice, complaint routes, enforcement, and correction. We keep recognition of a harm or benefit, tracking of outcomes, and recourse mechanisms. Niti Formalization and Nyaya Recourse are the framework's endpoints, and three components explain how formation settles the relation between them, drawing on Stone's polis~\cite{stone2012policy}, Lindblom's muddling through~\cite{lindblom1959science, lindblom1979still}, and Simon's bounded rationality~\cite{simon1955behavioral, simon2000bounded, march1993organizations}. Bounded rationality entered inductively, as first-wave participants described deciding with the knowledge, capacity, and time available, and it is the condition under which the other four operate. We adapt each theory to a condition it was not written to address (Table~\ref{tab:constructs}), and the result is an operational synthesis for studying policy in formation. 

\begin{table*}[t]
\caption{Components of the Society-Centered AI Governance Framework.}
\label{tab:framework}

\setlength{\tabcolsep}{4pt}
\begin{tabular}{>{\raggedright\arraybackslash}p{3.0cm}>{\raggedright\arraybackslash}p{4.3cm}>{\raggedright\arraybackslash}p{6.5cm}}
\toprule
\textbf{Concept} & \textbf{Framework component} & \textbf{Key governance question} \\
\midrule
\textit{Niti} (Sen) & Niti Formalization & What rules, inventories, and accountability mechanisms has the institution built? \\
\textit{Nyaya} (Sen) & Nyaya Recourse & What can a person learn, challenge, and have corrected? \\
\textit{Polis} (Stone) & Pluralistic Governance of AI & How do competing interests and actor power shape what gets built and who defines the problem? \\
Muddling through (Lindblom) & Adaptive Governance of AI with Incremental Revision & How does governance move forward as the technology and its consequences change? \\
Bounded rationality (Simon) & Bounded Rationality in AI Governance & What can institutions staff, interpret, enforce, and maintain, given limits of knowledge, capacity, and time? \\
\bottomrule
\end{tabular}
\end{table*}

The framework adds three claims to its source theories. One instrument can inform the state and give an affected person no way to respond, so visibility and contestability can separate inside a single document. A duty can bind an agency and leave the person affected without a right, so binding status alone cannot show protection. A commitment can pass to an instrument that does not yet exist, so the unit of analysis must be the sequence of instruments and the handoffs between them. The source theories explain why this happens: Sen sets the standard of what reaches people, Stone explains why some actors and harms enter governance earlier than others, Lindblom explains why correction proceeds through staged adjustment, Simon explains why institutions settle for arrangements they can staff, and Sanyal explains why plans anticipate institutional resistance~\cite{sanyal2005planning}. The framework tells an analyst what to read and what to count while AI policy forms.

\subsection{Procedural Visibility and Public Contestability}

An instrument has \textit{procedural visibility} when it tells the state which AI systems it runs, who is responsible for them, and how they were reviewed. It has \textit{public contestability} when an affected person can use it to learn that AI influenced a decision, challenge the decision, and see it corrected. Legal theory has long asked who holds the right that corresponds to a duty~\cite{hohfeld1917fundamental}, and we apply that question to AI instruments. An \textit{agency duty} runs through an institution's chain of command, while an \textit{affected-person right} is one the person can invoke, and public contestability requires the second.

\subsection{The Binding Sequence and Handoff Specification}

Governments often write authority, enforcement, and remedy into separate documents with different dates and legal force. Federal grantmaking shows this split ~\cite{bateyko2025one}, and North Carolina’s record shows it at the state level (Table \ref{tab:matrix}). The \textit{binding sequence} records, for each instrument, the date it was written and its binding status: in effect, enacted but not yet applicable, advisory, proposed, or absent. It also records whether a provision creates an agency duty or an affected-person right. \textit{Handoff specification} has two parts: \textit{implementation handoff} names who acts and by when, and \textit{enforcement handoff} names the later statute, budget line, agency rule, or contract clause that creates rights or duties.

A set of governance instruments exhibits visibility-first governance when the instruments that create procedural visibility for a class of AI systems have taken effect, and the instruments that would create public contestability for the same systems are advisory, proposed, absent, or in effect only as agency duties. The definition rests on binding status, and a legislature can end the pattern by enacting what is proposed. Supporting indicators apply these concepts (Table ~\ref{tab:constructs}): the \textit{standing gap} compares the publics a document names with the people on its roster, witness list, and drafting body, the \textit{agent test} asks which roles the causal story identifies before any remedy is written (a responsible institution, an intentional perpetrator, a regulated actor, an affected party), and \textit{capacity signatures} read limits of time, staff, and budget from a body's record (Table \ref{tab:capacity-subcodes}). Five further indicators complete the set (Table \ref{tab:constructs}). For Niti Formalization, the obligation rule counts a provision as formalization only when it creates a binding duty, which separates commitments from statements of intent. For Nyaya Recourse, preconditions of protection keep three levels separate: recognition of harms, affected groups, and benefit claims; tracking of outcomes; and recourse through complaints, pathways, enforcement, and correction. Depth of recourse specification then records the deepest level a document reaches for each named harm, whether the harm is named, tracked, routed to an office, or tied to enforcement and correction. For Adaptive Governance of AI with Incremental Revision, the revision test compares changes in institutional response with restated commitments, and the feedback condition asks whether a route exists for public consequences to reach decision-makers.

\subsection{Applying the Framework to Any Jurisdiction}
\label{sec:apply}

The framework asks for five steps wherever it is used: choose a class of AI systems, collect every instrument that governs it, record each instrument's date, binding status, and whether it creates an agency duty or an affected-person right, ask the four public questions, and trace each commitment's handoff. These steps need no outcome data and no particular legal system, so researchers can compare a national AI act, a state roadmap, and a city ordinance in the same terms. 

\subsection{Example Scenario: AI Data Center}
Consider a local decision to approve an AI data center. Niti Formalization records the zoning rules, utility agreements, disclosure requirements, and any advisory council that structures the decision. Nyaya Recourse asks whether the communities exposed to water use, electricity load, and utility costs can see those effects, challenge them, and seek correction. The center also brings jobs and tax revenue, so the public interest stays contested (Pluralistic Governance of AI). Officials decide under time pressure (Bounded Rationality in AI Governance) and revisit the decision as reports arrive (Adaptive Governance of AI with Incremental Revision). A disclosure rule may tell the state what a center consumes and leave residents with nowhere to complain.
\\ \\
\textbf{Membership} is a question the case must answer. The potentially affected publics, whether or not represented, include city residents and nearby communities, and the people who decide or advise include government officials who decide zoning and utilities, technology companies, environmental scientists, and environmental advocates. These actors form \textbf{groups}, such as neighborhood associations, environmental justice organizations, utility boards, planning departments, and the company's engineering team, and they hold unequal forms of \textbf{power}. An environmental advisory council can set disclosure requirements, impact assessment procedures, and oversight mechanisms \textbf{(Niti)}. Its roster shows who holds a seat, and the standing gap shows who does not.
\\ \\
The center brings construction jobs, long-term employment, tax revenue, and infrastructure investment. It also strains water and electricity, produces heat, and imposes environmental costs on communities that did not choose to host it. These tensions are opportunities for \textbf{common interests}. The company helps define acceptable water and energy use, and community groups use \textbf{community and collective will} to contest that definition. Distinct \textbf{motivations} pull governance in different directions: the company wants to build for profit, the city government sees the \textbf{market logic} of investment and jobs, and residents and advocates press for safety and sustainable practices. \textbf{Public interest} emerges as a set of contested questions about the center's purpose, who bears its costs, and who benefits.
\\ \\
Members draw on scientific studies, sustainability reports, advocacy campaigns, public testimony, and independent audits, and this \textbf{information} comes from actors who exert \textbf{influence} over how the problem is framed. Whether the council's decisions reduce harm for the communities most exposed is the \textbf{Nyaya} question. Decisions satisfice under \textbf{bounded rationality}, \textbf{co-operation} carries the process forward while motivations conflict, \textbf{loyalty} sustains the council across meetings, and the council adapts as technology and its consequences change \textbf{(muddling through)}. The framework applies whether structures are being built or revised.
\\ \\
Council members raised this case themselves. One said that \textit{``when we talk about AI, we are talking about an energy and a land use discussion''} (P8), and another recalled, \textit{``People were scared of them ... they didn't want their energy costs to go up''} (P7). One member doubted that proposed disclosure would measure the harm: \textit{``Requiring disclosure of token consumption is one indirect indicator of energy use, but if you're not mandating that they also report on their energy source, then that token number is fairly meaningless.''} A third member proposed a negotiated exchange, a form of community benefits agreement: \textit{``What if you said: you can put it in my backyard, but I want 10\% of all the tokens for free''} (P3).

\section{Study Design: A Temporal Case of AI Policy in Formation}
\label{sec:method}

To understand how AI policy forms and how officials decide, we follow one state before, at, and after the release of its AI plan. Our design is abductive~\cite{timmermans2012theory} and links theory, practitioner interviews, and document analysis in one workflow. We built a codebook from Sen, Stone, and Lindblom (Table \ref{app:codebooks}). First-wave interviews revised it by adding bounded rationality, we applied the revised codebook to the documents, and second-wave interviews checked the components against accounts of the released Roadmap. The methods complement one another: AI governance documents record what was decided and rarely record the limits under which officials decided it, so bounded rationality came from the interviews, and the model-assisted pass then let us look for its signatures across the AI governance corpus.

\paragraph{AI governance corpus.} The AI governance corpus contains 49 documents published between 2020 and 2026, totaling about 475,575 tokens (Appendix~\ref{app:document-corpus}), and we use the document set and selection criteria from prior work~\cite{migineishvili2026regulating}. The 31 North Carolina documents address AI governance. Twenty-nine come from official state bodies (17 executive, 10 legislative, and 2 judicial). Two come from outsider publishers and serve as contextual sources: an AI Center for Government article on a state program (NC, ID-13) and a public college School of Government post on local court orders (NC, ID-29). 

\paragraph{Semi-structured interviews.} We wanted to understand Council-level reasoning while the Roadmap formed, so we used purposive sampling~\cite{palinkas2015purposeful} to select people with direct involvement in that process. From public membership records, we selected nine Council members covering all four subcommittees. We judged adequacy by information power~\cite{malterud2016sample}: the aim is narrow, participants hold specific knowledge of one governance process, established theory guided the analysis, and we read the interviews alongside public documents and a provision-level reading. The first wave (P1--P5) took place in April and June 2026, before the Roadmap's release, and the second (P6--P9) between July and September 8, 2026, after it. The last interview took place on September 8, the same day as our observation cutoff. The waves involved different participants, so we analyze reasoning at the Council level. Interviews lasted 30 to 45 minutes, and we recorded them after verbal consent under IRB approval and wrote field notes after each session. Participants could withdraw at any time, and only the team could access recordings and transcripts. To protect anonymity, we use pseudonyms, omit roles and subcommittees, and leave the pseudonym off the most specific accounts.

\paragraph{Model-assisted coding.} We limited model assistance to the public AI governance corpus and excluded interview transcripts. GPT-5.4 generated candidate inductive labels independent of our framework and candidate assignments to its five components, and GPT-4o-mini proposed mappings between the two (Appendix~\ref{app:prompts}, which also gives settings and prompts). Table \ref{tab:workflow} summarizes the workflow. Following work on text as data~\cite{grimmer2013text}, we treated every output as a candidate to validate, and all claims about binding status and legal force come from the human planning reading. All 49 documents were coded in a single pass on July 7, 2026, and the document set did not change afterward (Appendix~\ref{app:document-corpus}).

\paragraph{Planning reading and interview analysis.} A planning reading examines instruments for their order and legal force. We counted an arrangement as formalization only when it creates a duty, and recorded each instrument's date and binding status as of September 8, 2026, our observation cutoff, using the General Assembly's bill records, and placed each named harm by depth of recourse specification (Tables ~\ref{tab:depth} and ~\ref{tab:constructs}). Binding status of the document can change while text does not, so this cutoff is separate from the July 7 model-assisted coding of document text. Coding cannot show that a decisive provision was not missed, so we also read the instruments behind Tables ~\ref{tab:matrix} and~\ref{tab:depth} for edition, scope, exceptions, and any notice, challenge, or correction provision. Two other executive instruments, a security policy on high-risk applications (NC, ID-10) and a description of how use cases are evaluated (NC, ID-11), concern devices, networks, and evaluation procedure rather than notice, challenge, or correction, so Table 2 omits them. A negative classification means that we identified no such provision in the corpus. Two researchers applied the codebook to transcripts in Taguette~\cite{Rampin2021} and used reflexive thematic analysis~\cite{braun2006using, braun2019reflecting}. The team also developed 13 inductive interview codes (Table \ref{tab:inductive-codebook}) and ran a reliability check on their excerpt-level application (Cohen's $\kappa = 0.812$).

\paragraph{Human review.} A researcher checked every retained model assignment against its source. Two researchers each coded 10 random documents for the five components (100 binary decisions), which matched the model’s on 97.0\% (Cohen's $\kappa = 0.808$). We report this as validation of model output, separate from intercoder reliability. The random sample also checks for missed assignments, though exhaustive recall would need a separate measure. Through discussion, reviewers resolved 77 unmatched inductive assignments by function (Appendix~\ref{app:inductive-mapping}), including passages on staffing limits and legal uncertainty, which they assigned to Bounded Rationality in AI Governance. The final dataset contains 8{,}589 mapped code instances; Table \ref{sec:complement} lists what each analysis contributed.

\section{Findings}
\label{sec:findings}

\subsection{Visibility-First Governance for the State's Own AI Systems}
\label{sec:sequence}

Procedural visibility took effect through executive action. The Responsible Use Framework requires agencies to maintain inventories and to monitor, audit, and document compliance (NC, ID-6). Executive Order No.~24, signed on September 2, 2025, directed that \textit{``each Cabinet agency shall establish an Agency AI Oversight Team''} and provided for high-risk use cases to be collected and made available for public review (NC, ID-1; ID-9). The Roadmap, released in July 2026, builds on these steps with 25 recommendations under 17 goals (NC, ID-49).

\begin{table*}[t]
\caption{Provision-level reading of the corpus instruments that govern North Carolina's own AI systems, in date order, with binding status as of September 8, 2026. ``None'' means that we identified no AI-specific provision in the instrument. ``Framework'' is the Responsible Use Framework. Dates in the Instrument column are issue dates; dates in other columns are Roadmap timelines.}
\label{tab:matrix}

\setlength{\tabcolsep}{3pt}
\begin{tabular}{>{\raggedright\arraybackslash}p{2.6cm}>{\raggedright\arraybackslash}p{2.6cm}>{\raggedright\arraybackslash}p{2.0cm}>{\raggedright\arraybackslash}p{1.8cm}>{\raggedright\arraybackslash}p{1.8cm}>{\raggedright\arraybackslash}p{1.6cm}>{\raggedright\arraybackslash}p{1.6cm}}
\toprule
\textbf{Instrument} & \textbf{Visibility duty} & \textbf{Notice} & \textbf{Challenge} & \textbf{Correction} & \textbf{Later instrument} & \textbf{Status} \\
\midrule
Framework v1.1 (Aug 2025; ID-6) & Inventories; monitoring, audit, and documentation & Agency ``shall provide notice''; disclaims enforceable rights & None & None & Roadmap; no right named & In effect as internal direction \\
\addlinespace[1pt]
Executive Order No.~24 (Sep 2025; ID-1) & Oversight teams; high-risk use cases collected for public review & None & None & None & Roadmap; no right named & In effect \\
\addlinespace[1pt]
Roadmap, agency AI generally (Jul 2026; ID-49) & Recommends public dashboards of high-risk use cases & As Framework & Recommends feedback mechanism (Dec 2026) & Recommends AI Liaisons for escalation (Dec 2027) & Owner and date named; no instrument named & Advisory \\
\addlinespace[1pt]
Roadmap, benefits eligibility (Jul 2026; ID-49) & As above & Recommends disclosure (Mar 2027) & None; general appeal only & Person keeps final determination & Owner named; no instrument named & Advisory \\
\bottomrule
\end{tabular}
\end{table*}

Notice is the closest case. The Responsible Use Framework states that \textit{``the user agency shall provide notice to those who may be impacted by its use,''} and it also states that it does not \textit{``create any right or benefit, substantive or procedural, enforceable''} against the State (NC, ID-6), so notice is an agency duty without an affected-person right.

Challenge and correction sit further ahead, and the Council can only advise: \textit{``the committee, frankly, can't even make that kind of a decision in an official capacity''} (P1). The Roadmap states the concern in broad terms: its principles call for people to \textit{``challenge outcomes when necessary,''} its transparency goal promises \textit{``clear pathways to raise concerns,''} and its feedback mechanism would \textit{``measure public sentiment and improve tools based on input''} (NC, ID-49). Its commitment that AI systems \textit{``do not make final benefits determinations''} (NC, ID-49) is a human-decision safeguard, which keeps a person responsible (\textit{``We're not going to take people out of the process,''} P7) and gives a claimant no new route to complain. We identified no instrument in effect in the corpus that creates an AI-specific right to challenge or correct an AI-enabled decision by the state.

\paragraph{General and AI-specific recourse.} A claimant may already hold rights outside these instruments: public assistance applicants have a statutory appeal through a county hearing, a state hearing, and judicial review~\cite{ncgs108a79}, and due process and civil rights law also apply. The state's Department of Justice applies consumer protection law to AI practices (NC, ID-49). A claimant can review the case file on appeal~\cite{ncgs108a79}, and a person who does not know AI was involved has little reason to raise its role. Our claim concerns whether AI instruments add notice a person can invoke, an AI-specific challenge, or correction.

After the Roadmap's release, some participants placed the binding step with the budget and the legislature: \textit{``The strategic roadmap has recommendations, the budget has mandates''} (P6). The record is a documented, incomplete sequence, since the Roadmap dates its minimum-standards recommendation September 2026 and its feedback mechanism, benefits disclosure, and AI Liaisons later, so the corpus cannot show whether they took effect.

\subsection{Why Visibility Took Effect First: Authority, Capacity, and Decision Logics}
\label{sec:shape}

\subsubsection{Divided Authority and the Legislative Path}
\label{sec:why}

Executive authority can create inventories and oversight teams alone, and public enforcement and a private right of action require the General Assembly. House Bill 934 (NC, ID-19) and Senate Bill 735 (NC, ID-25), two corpus bills with enforcement provisions, remain in committee, and legislators have voted neither down. House Bill 934 received a favorable report from Regulatory Reform on April 29, 2025 and a favorable committee substitute from Judiciary 3 on May 6, then went to Election Law, where it remains. The Senate moved Senate Bill 735 from Rules to Appropriations/Base Budget on March 26, 2025, the day after filing. Participants’ accounts, which we coded as multilevel government authority, describe the same arrangement. At least one member expected limited legislative action in regards to the creation of the NC AI Leadership Council: \textit{``there's probably not going to be legislative action on this...it sort of laid the foundation for we need an executive order, we need an AI leadership council, we need to create accelerators inside the different departments''} (P2). 

The two paths also face different tests. A statute must withstand legal challenge, including federal challenge to state regulation (P4), and one member counted litigation among \textit{“the two big challenges”} for AI (P8). Another recalled difficulty defining high-risk data. An executive order that directs agencies to inventory their own systems faces fewer legal and drafting tests than a recourse statute.

Some members also doubted that general rules could fit the technology: \textit{``It has to be grounded in a specific use case''} (P1). One member expected that \textit{``most of the meat-and-potatoes regulatory decisions ... they're going to get made in the agencies that already regulate those spaces''} (P1). This expectation makes the Council’s advisory role deliberate.

\subsubsection{Institutional Capacity and Symbolic Dynamics}
\label{sec:capacity}

Another reading treats state AI governance as symbolic, driven by growth ambitions or shaped by vendors~\cite{bietti2020ethics, hagendorff2020ethics}. Institutional theory holds that organizations can adopt legitimating structures in good faith~\cite{meyer1977institutionalized, edelman1992legal}. Participants' accounts cut both ways: a member described industry as \textit{``begging for guardrails''} (P8), while another warned that the large model providers have \textit{``enormous influence''} and \textit{``their own interests''} (P2).

Capacity was a recurring explanation among the members we interviewed: one said that downstream implementers were overwhelmed and that some commitments were paced to available capacity because they \textit{``would require too much effort and human capacity from the people meant to fill them''} (P6). Symbolic dynamics can operate alongside this account, since capacity limits, growth politics, and industry influence can work together.

An inventory fits an existing office and reporting routine~\cite{cyert1963behavioral, nelson1985evolutionary}, while a complaint process with authority needs new staff and legal standing. States also reused each other's tools, and inventories recur in Oklahoma and Texas (OK, ID-42; TX, ID-45). The Roadmap says that \textit{``the State will need to realign its existing funding''} (NC, ID-49).

\subsubsection{Decision Logics Behind the Roadmap's Form}
\label{sec:form}

The Roadmap is broad, near-term, and advisory, and the members we interviewed explained this form through four decision logics. The first treated AI as unavoidable (\textit{``it doesn't matter if you like it or not, it's coming either way,''} P8), the second was competitive timing (\textit{``the first states to organize are the ones who are going to win,''} P7), the third was uncertainty (\textit{``We're not putting out a 10-year roadmap for AI,''} P5), and the fourth made trust the starting point: \textit{``Everybody felt that way: we had to have trust to move forward. And you saw trust become a framework for policy''} (P7). These logics favored instruments that the executive could adopt quickly and that the public could see, while binding recourse fit them less easily because it required definitions and statutory authority and its value would appear mainly after a harm occurred.

\subsection{Handoff After the Roadmap}
\label{sec:advisory}

The Roadmap has strong implementation handoff: every recommendation shows a primary owner and a completion timeline, and the Council will report again in June 2027 (NC, ID-49). Table \ref{tab:roadmap} lists selected recommendations. Enforcement handoff differs by harm. For societal harms, the Roadmap names a vehicle and a time, a policy package on child safety, privacy, and fraud \textit{``to be advanced in the 2027 legislative cycle''} (NC, ID-49), and for the state's own systems, the provisions we examined name no statute, budget line, agency rule, or contract clause that would give an affected person a right. A plan can exist before a bill number does, so the gap we report concerns enforcement specification, which a provision-level handoff would close by mapping each recommendation to the provision that carries it. One member described a shift \textit{``from planning and writing to execution''} and named the intended scope of integration as \textit{``across procurement, cybersecurity, 911, broadband,''} while noting that the state was \textit{``not quite where we need to be yet.''}

\subsection{The Distribution of Recourse Across Harms: Remedies and Standing}
\label{sec:reach}

\subsubsection{Causal Stories and the Allocation of Remedies}
\label{sec:remedies}

Our analysis shows how recourse is distributed, and its instruments do not establish a sequence for the state's own systems. The documents name harms directed at children most directly. California warns of algorithmic error across housing, education, employment, credit, health care, and criminal justice (CA, ID-32), and Washington's report links AI governance to sextortion, hiring errors, and banking discrimination (WA, ID-47). Recognition of harms and groups is the most specified level, and the mechanisms that would measure harms, route complaints, enforce rules, or require correction appear with less procedural detail. Read from top to bottom, Table~\ref{tab:depth} shows specification falling as the responsible party moves from individual offenders and developers to deploying agencies and program owners.

\begin{table*}[t]
\caption{Depth of recourse specification for harms named in the North Carolina record, with binding status as of September 8, 2026.}
\label{tab:depth}

\setlength{\tabcolsep}{4pt}
\begin{tabular}{>{\raggedright\arraybackslash}p{3.8cm}>{\raggedright\arraybackslash}p{4.5cm}>{\raggedright\arraybackslash}p{2.4cm}>{\raggedright\arraybackslash}p{1.2cm}>{\raggedright\arraybackslash}p{3cm}}
\toprule
\textbf{Harm} & \textbf{Deepest level specified} & \textbf{Instrument} & \textbf{Status} & \textbf{Responsible party} \\
\midrule
AI-generated sexual imagery of minors; altered private images & Enforcement: felony offenses; civil action and an order destroying the image & Session Law 2024-37 (NC, ID-21) & In effect & Individual offender \\
Deepfake harms to individuals & Enforcement: offense, civil cause of action, and image destruction & House Bill 934 (NC, ID-19) & Proposed & Individual creator or distributor \\
Errors when professionals use AI & Liability allocated: professional solely responsible; developer shielded & House Bill 934 (NC, ID-19) & Proposed & Learned professional \\
AI safety incidents & Tracked and routed: 72-hour report to the attorney general, with enforcement authority & Senate Bill 735 (NC, ID-25) & Proposed & Developer \\
AI in benefits decisions & Human-decision safeguard: no final determinations by AI; no complaint route & Roadmap (NC, ID-49) & Advisory & Deploying agency \\
Public concerns about agency AI use & Routed to staff: public feedback mechanism & Roadmap (NC, ID-49) & Advisory & Deploying agency \\
Unequal access, workforce displacement & Programmatic response; no individual complaint or correction right & Roadmap (NC, ID-49) & Advisory & Program owners \\
\bottomrule
\end{tabular}
\end{table*}

Enacted and proposed enforcement provisions address individual offenders, creators, and developers, and one of them is in effect. Session Law 2024-37 makes the creation, distribution, and possession of AI-generated sexual imagery of identifiable minors a felony (NC, ID-21), and it extends the private-images statute, under which a person depicted can sue for damages and an order destroying the image, to realistic depictions created with AI~\cite{ncgs14190}. House Bill 934, filed on April 10, 2025, would give \textit{``any person aggrieved by a violation''} (NC, ID-19) of its deepfake provisions a civil cause of action. It would also shield the developer of an AI product from liability for errors when a learned professional uses it and make that professional solely responsible to the client, so the bill allocates liability as well as remedies (NC, ID-19). Senate Bill 735 creates no general private right of action (NC, ID-25), and neither bill has moved since 2025 (Appendix~\ref{app:evidence}). For harms from the state's own use of AI, the responsible party is the deploying agency, and the Roadmap offers a human-decision safeguard and a staff-read feedback channel.

Enforcement clauses must name whom they bind, so the agent test reads the story told before any remedy. Stone argues that stories with an intentional cause are the easiest to turn into policy~\cite{stone2012policy}. The Roadmap's background on societal harms has this form: it reports victim complaints and losses and names the elderly and children as those most harmed. Its benefits recommendation describes a different risk, that AI use could \textit{``compromise the integrity of benefits eligibility determinations''} (NC, ID-49), and it names no victim and no wrongdoer. The deploying agency in a benefits decision is identifiable, and Senate Bill 735 regulates developers without alleging intent (NC, ID-25), so neither identification nor intent alone separates these cases, and we report this as a descriptive pattern.

AI governance research usually treats harm as something a system does to a person, through error or a deepfake, and participants also described a harm of exclusion. Unequal access is the diffuse harm they discussed most, with concern for rural communities, students, seniors, low-income residents, and small businesses. One member described the aim of the work as \textit{``how you don't make the digital divide even worse''} (P3), another explained why infrastructure alone would not be enough (\textit{``I could run fiber straight up to your front porch, but if you can't afford the internet, or don't have a laptop or tablet ... you could still get left behind,''} P9), and a third set an order for policy: \textit{``We shouldn't spend a dollar training someone to use something that everyone doesn't have equal access to''} (P8). Exclusion has no perpetrator and no single aggrieved person, so the Roadmap responds with programs such as literacy training and compute access. Some participants connected rural access to broadband, public libraries, training, and public services, and the comparative documents usually folded access into broader categories.

\subsubsection{Standing and Representation in Deliberation}
\label{sec:standing}
Standing decides who tells these stories. Georgia's study committee describes a public of citizens, students, workers, and patients, and its members are senators alongside academic and industry experts (GA, ID-35). Washington's Attorney General's Office acknowledged that the Legislature had not included Tribes in the task force membership and added supplemental advisory structures (WA, ID-47). Kentucky's task force heard from Microsoft, Google, Amazon, and Salesforce, among others (KY, ID-38). North Carolina's Roadmap credits stakeholders \textit{``across government, private sector, education, and labor''} (NC, ID-49), and some members raised the same concern themselves. \textit{``The big gap is the average North Carolinian,''} one said (P2). Another listed who was absent: \textit{``there were no community voices, no activists, no one who might represent communities that could be opposed to or disproportionately affected by data centers being sited in their areas,''} and \textit{``youth were not included''} (P4). 
Vendor contact exceeded constituent contact for one member: \textit{``I hear feedback more from ... a vendor point of contact than a constituent one''} (P8). We have no record of who advocated for Session Law 2024-37 or House Bill 934, so we present standing as a condition that co-occurs with the pattern. Appendix \ref{app:evidence} gives additional evidence on remedies and standing.

For a benefits claimant in North Carolina, the state can see the system in its inventory, the agency holds a duty under the Responsible Use Framework to give notice that the claimant cannot enforce, the Roadmap recommends disclosure and a person at the final determination, and the general public assistance appeal remains. We identified no instrument in effect in the corpus that gives that claimant an AI-specific right to learn AI was used, challenge its role, or have an AI-related error corrected.

\section{North Carolina Among the States}
\label{sec:shared}

North Carolina's record has several features that the framework credits. It has instruments across executive, legislative, and judicial contexts and an enacted AI-specific criminal law with a civil remedy. Every Roadmap recommendation has a primary owner and a completion timeline, and the Council will report publicly again in June 2027. The Roadmap recommends public dashboards of high-risk use cases, disclosure of AI use in benefits eligibility, dedicated AI Liaisons in each Cabinet agency, and third-party monitoring of data center impacts (NC, ID-49). North Carolina's record is fuller in our dataset because our design includes multiple instrument types for the state, whereas the comparative corpus includes one report per state.

The comparative reports share tool vocabularies and differ in which bodies hold authority. Participants described the same search: \textit{``Who else has done this? What other state has already started and tried something on this?''} (P2). California sorted oversight by risk tier (CA, ID-32), Oklahoma paired a proposed Chief AI Officer with annual agency inventories (OK, ID-42), and Texas recommended comprehensive inventories and mandatory risk assessments for high-risk deployment (TX, ID-45). Several states divided the work into parts, with four working groups in Illinois (IL, ID-36), eight subcommittees under Washington's statute (WA, ID-47), and four subcommittees on North Carolina's Council (NC, ID-5). One member described staff as important for \textit{``desk research of what other states have done''} and said that the four subcommittees reflected \textit{``some of that preliminary work, trying to carve up the universe''} (P1). California and Arkansas used the NIST AI Risk Management Framework, a voluntary framework written for organizations managing their own risk~\cite{nist2023airmf} (CA, ID-32; AR, ID-31). The comparative reports show similar tool vocabularies and recurring emphasis on inventories and risk tiers, and each state's binding sequence would require its other instruments.

AI governance settlements vary across jurisdictions. New Jersey's approach relies on small pilots under a test-and-learn model (NJ, ID-40). Comparative work finds that AI governance documents allocate authority and political meaning differently across jurisdictions~\cite{bareis2022talking, bogiatzis2024beyond, schiff2023looking}, and Pluralistic Governance of AI explains this divergence, since local politics decides who takes part and which harms they raise.

Much AI governance research codes each document for what it contains~\cite{migineishvili2026regulating, alanoca2025comparing}, which gives low marks to an advisory roadmap without enforcement provisions. Open delegation is common across the comparative reports. Vermont's task force proposed a code of ethics and a permanent oversight commission as a first step toward considering regulation (VT, ID-46), and Illinois and California each described their reports as first steps in an iterative process (IL, ID-36; CA, ID-32). Each document leaves details to later instruments, and few specify what those instruments must contain. Federal AI programs also delegate implementation to agencies and lower-level offices~\cite{bateyko2025one}. Vermont's task force recorded that the time available was too limited to determine which AI applications agencies should use (VT, ID-46), and Connecticut recommended focusing resources on the systems with the highest impact on individual rights and called the approach pragmatic (CT, ID-34).

\section{Discussion}
\label{sec:discussion}

AI governance is usually judged by what it has produced, and our findings suggest a second test: what a person can do with it. North Carolina's record is strong on the first test and shows what remains open on the second, and the same may hold for other governments that acted early. The question matters now because instruments written in this period set what later evaluation can measure, whose harms count, and which duties later governments inherit.

\subsection{The Findings in Research and Policy Debates}
\label{sec:debates}

Three results carry beyond this case: a duty on an agency can exist without a right a person can invoke, the instruments in effect concern what the state can see, and the one enacted remedy addresses a harm with an intentional offender and a named victim, so the harms that reach enforcement first may differ from the harms of a government's own AI use.

These results change what counts as progress and compliance. A government can add AI laws, inventories, and oversight teams without adding anything a resident can use, and an agency can meet every internal duty while the person affected never learns that AI was used. They extend three lines of research: audit work~\cite{raji2020closing, groves2024auditing} gains the question of who can invoke a standard, contestability research~\cite{vaccaro2019contestability, lyons2022s, alfrink2023contestable} gains the policy instruments that create its routes, and participation research~\cite{corbett2023power, birhane2022power} gains the standing gap.

The findings also bear on two policy debates. Federal preemption remains open~\cite{ropes2026limits, casrai2026preemption}, and states can act on the systems we study however it is resolved, so the question of who can invoke a duty stays within a state's reach. On enforcement design, Colorado's replacement law gives consumers notice, correction, and human review, enforced by the attorney general with no private right of action~\cite{eckert2026colorado, colorado_sb26_189}, while House Bill 934 would create a civil action. Agency duties, rights a public enforcer protects, and rights a person can take to court form a range, and drafters can use the four public questions before adoption.

\subsection{Legitimacy Before Protection}
\label{sec:legitimacy}

Three readings fit the record so far. The first is planned sequencing, for which the Roadmap's owners, dates, and June 2027 report are evidence, and the second is deferral, in which each first step points to another and no binding instrument follows. The reviewed record does not yet establish which successor instruments will create AI-specific public rights.

The third reading holds that visibility produces something of its own. One member described the aim in terms of protection: \textit{``We started with trust, making sure every North Carolinian felt protected''} (P7). The Roadmap's letter calls the work \textit{``an investment in the public trust''} (NC, ID-49), and some participants made trust the starting point of their reasoning (Section~\ref{sec:form}). If visibility increases trust while AI-specific contestability remains advisory or dependent on later implementation, early AI governance may produce legitimacy before protection, and a legislature that sees inventories and a published roadmap may treat AI as governed. We identify this as a testable risk of visibility-first governance: we have not shown that it occurs, and we make no claim that officials intended to substitute trust for protection. The stakes reach beyond one state, because the Roadmap reports that several states use disclosure to strengthen public trust (NC, ID-49).

The Roadmap's own dates can separate these readings. Planned sequencing predicts that the December 2026 feedback mechanism names who receives a concern and what that office can do, that the March 2027 benefits disclosure reaches claimants and links to an appeal, that the June 2027 report maps recommendations to provisions, and that the 2027 session moves bills that match each recommendation. Deferral predicts restated commitments at those dates, and legitimacy before protection predicts that officials cite dashboards or the Roadmap as evidence of protection or as a reason to wait.

\subsection{The Durability of Visibility and Recourse}
\label{sec:beyond}

The framework reads public-record cases beyond our AI governance corpus in the same terms, and they show how long each instrument lasts. The European Union placed recourse duties in legislation~\cite{eu2024aiact}, and an amending regulation deferred the obligations for stand-alone high-risk systems to December 2, 2027~\cite{eu2026omnibus}. Canada's recourse rests on a Treasury Board directive the executive can amend, and the Artificial Intelligence and Data Act lapsed in January 2025~\cite{canada2019dadm, mcinnes2025aida}. Executive visibility was fast and persisted across administrations \cite{eo13960}, administrative recourse was fast and revocable, and statutory recourse was durable and slow, so the form of an instrument decides how long a protection lasts. North Carolina's notice duty sits in an executive policy, which places it in the second category.

\subsection{Qualifying Planning and Policy Theory for AI Governance}
\label{sec:heldup}

The case qualifies the framework's source theories in three places. First, implementation structures took effect by executive order without a statute~\cite{sabatier1986top, hupe2014happens}, and an agency duty took effect that no person can enforce, so compliance alone cannot show that the public gained a right. Second, Stone's polis describes a contest among actors~\cite{stone2012policy}, and our case adds that some harms never enter it, because the publics they affect are absent from rosters and testimony lists, and Davidoff's advocates~\cite{davidoff1965advocacy} have no counterpart here. Third, Simon's bounded rationality usually treats capacity as a limit on choice~\cite{simon1955behavioral, march1993organizations}, and our case adds that budgets allocate it.

Incremental revision needs the largest qualification. One member described guidance as \textit{``a living document''} (P6) and the Roadmap added advisory commitments across many domains (NC, ID-1; ID-49), most of which do not create individual complaint, review, or correction rights. The case partly meets Lindblom's assumption that decision-makers learn from each step~\cite{lindblom1959science}, and we read incremental revision as conditional on recourse, because recourse determines whose experience the next increment responds to.

\subsection{Implications for Policymakers, Researchers, and the Public}
\label{sec:practice}

Each implication follows from a finding and needs no outcome data, and policymakers at state, federal, and international levels can adopt six practices.
\begin{itemize}
  \item \textbf{Test each instrument at drafting.} Ask the four public questions of a bill, order, memorandum, roadmap, or international agreement before adoption, since an order can establish oversight and leave the complaint route to a later instrument (Table ~\ref{tab:fourquestions} in Appendix E.1).
  \item \textbf{Name the enforcing instrument.} State which statute, rule, budget line, or contract clause will enforce each recommendation and who can invoke it. North Carolina's Roadmap already assigns an owner and date to all 25 recommendations, so this is a natural next step (Section~\ref{sec:advisory}).
  \item \textbf{Turn agency duties into rights where the harm warrants it.} Let an affected person enforce a duty through a complaint route or a civil action. North Carolina's notice duty could become enforceable through its 2027 session or an agency rule, with Colorado's law as one model, and the federal removal of notice in 2025 shows why the form of a duty matters (Table~\ref{tab:matrix}).
  \item \textbf{Sort remedies by responsible party.} List the party responsible for each harm and designate an agency to answer for diffuse harms. Across seven harms in our record, enforcement reaches offenders, creators, developers, and deploying agencies answer through recommendations (Table~\ref{tab:depth}).
  \item \textbf{Link the records.} Hold the affected groups and appeal route in the inventory, and tie complaints to a system with a shared identifier. A complaint needs a system to name, and public inventories exist at federal and state levels: Maryland lists 75 use cases and Connecticut 31 systems (NC, ID-49).
  \item \textbf{Give feedback to an owner.} State who may submit a concern, which office receives it, its authority, and when a reply is due, starting from North Carolina's planned mechanism.
\end{itemize}

Researchers can use the five steps (Section~\ref{sec:apply}) and a provision-level reading to compare jurisdictions, and they can score the Roadmap's dates against the three readings. The public can ask the four questions of any published plan without reading a statute: residents, journalists, and advocates can ask where a person can complain and who can correct a decision, and an unanswered question is itself evidence of what remains open.

\subsection{Scope Conditions and Limitations}

Our finding describes North Carolina, and the framework is what we generalize. We expect visibility-first governance where an executive can create inventories and oversight alone, binding rights require a legislature, and the body writing the plan holds advisory authority. The pattern should weaken where a legislature acts first, as in the European Union and Colorado, whose statutes carry recourse duties. The finding also holds as of our cutoff, since many Roadmap dates have not arrived, and the Roadmap's milestones would show whether it lasts. Our analysis covers the AI governance corpus and the interviews, so bills and laws outside the corpus, including House Bill 1161~\cite{nc2026h1161} and the 2026 Appropriations Act~\cite{nc2026s257summary}, may change the status of a class of systems. General appeals may already protect people, so our claim concerns what AI instruments add, and an appeal is hard to use without notice that AI was involved. Agency duties can also work without private rights, so we ask who can act when oversight does not.

We identify institutional preconditions for public protection, and further study must show whether particular systems protected or harmed people. The 18 comparative reports in the AI governance corpus let us compare patterns (Appendix~\ref{sec:shared}), and we cannot use them as a census of state AI governance or a ranking of state performance. Each state appears through one report, so they cannot show its binding sequence. Nine Council members gave us information-rich accounts of one process, and we did not hear from residents who live with these systems. Future work can test these conditions in other jurisdictions, measure lived outcomes with affected residents, workers, educators, advocates, agency staff, procurement officials, and vendors, and examine whether inventories, feedback channels, oversight teams, procurement checkpoints, and revision logs make AI use more visible, contestable, and correctable.

\section{Conclusion} 
A government can see its AI before the people affected by it can do anything about it. That distance is easy to miss, because an inventory, a public roadmap, and even a duty to give notice can exist without a right a person can invoke. North Carolina moved early, each branch acted within its authority, and its Roadmap sets dates for the next steps, so the state can still close that distance. Every government now writing AI policy decides how wide that distance will be, because its instruments shape who can see a system, who can contest it, and who can have it corrected. The Society-Centered AI Governance Framework gives policymakers, researchers, and the public in any jurisdiction a shared language for that decision before outcomes exist, and its test is who can act on what an instrument makes visible.

\section*{Ethical Considerations Statement}

This study received IRB approval, and participants gave verbal consent and could withdraw at any time. Council rosters are public, so we use pseudonyms, omit roles and subcommittees, and leave the pseudonym off the most specific accounts. We analyze the institutional conditions that shape visibility, contestability, and recourse, and we make no claim about the intent or fault of any individual or agency.

\section*{Generative AI Usage Statement}
We built the codebook from theory and used GPT-5.4 and GPT-4o-mini on public documents, with human validation, as Section~\ref{sec:method} and Appendix~\ref{app:prompts} describe. Claude Opus 5.5 was used for grammar and language editing of author text, as per conference guidelines.

\section*{Acknowledgments}
We thank the UNC AI Acceleration Program for providing compute resources and study participants for their valuable time and contributions.

\bibliographystyle{ACM-Reference-Format}
\bibliography{sample-base}

\newpage
\appendix

\onecolumn 

\section{AI Governance Corpus}
\label{app:method-details}
\label{app:data}

This appendix lists the 49 documents in the AI governance corpus. We treated model outputs as candidate annotations, and human researchers made the final analytic decisions.

\subsection{AI Governance Corpus}
\label{app:document-corpus}

Table~\ref{tab:document-corpus} lists the 49 documents. Twenty-nine of the 31 North Carolina documents come from official state bodies (17 executive, 10 legislative, 2 judicial). The other two come from outside publishers and serve as contextual sources: an AI Center for Government article on a state program (ID-13) and a UNC School of Government post on local court orders (ID-29). The 18 comparative documents are reports from state task forces, committees, and agencies, one per state. Table~\ref{tab:document-provenance} gives each document's issuing body, issue date, version or bill edition, and the date we retrieved it. We retrieved the Roadmap (ID-49) on July 1, 2026 and every other document on May 16, 2026. We coded all 49 documents on July 7, 2026, after retrieval, and did not add, replace, or remove any documents afterward. Binding status was recorded separately as of September 8, 2026 and did not change the document set. Bills marked edition 0 (IDs 22--27) are pre-filing drafts. We read the draft text at the linked General Assembly address and confirmed each bill's number, session, and filing status from the General Assembly's bill records as of our September 8, 2026 observation cutoff.

\begin{longtable}{@{}L{0.8cm}L{1.0cm}L{1.9cm}L{2.3cm}L{7.7cm}@{}}
\caption{AI governance corpus. Document identifiers are internal archive IDs. IDs 13 and 29 come from outside publishers and are listed separately as contextual sources.}
\label{tab:document-corpus}\\
\toprule
\textbf{ID} & \textbf{State} & \textbf{Branch} & \textbf{Type} & \textbf{Document title} \\
\midrule
\endfirsthead

\caption[]{AI governance corpus. Continued.}\\
\toprule
\textbf{ID} & \textbf{State} & \textbf{Branch} & \textbf{Type} & \textbf{Document title} \\
\midrule
\endhead

\midrule
\multicolumn{5}{r}{\emph{Continued on next page}}\\
\endfoot

\bottomrule
\endlastfoot

\multicolumn{5}{@{}l}{\textbf{North Carolina: Executive Branch}}\\
\addlinespace

1  & NC & Executive & Executive order & Executive Order No.\ 24: Advancing Trustworthy AI That Benefits All North Carolinians \\
2  & NC & Executive & Press release & Governor Stein Announces Executive Order on AI \\
3  & NC & Executive & Membership roster & AI Leadership Council Members \\
4  & NC & Executive & Meeting minutes & 11/05/25 AI Leadership Council Meeting Minutes \\
5  & NC & Executive & Meeting minutes & 01/21/26 AI Leadership Council Meeting Minutes \\
6  & NC & Executive & Governance framework & NC State Government Responsible Use of AI Framework \\
7  & NC & Executive & Governance principles & Principles for Responsible Use of AI \\
8  & NC & Executive & Use policy & Use of Publicly Available Generative AI \\
9  & NC & Executive & Inventory & AI Inventory \& Assessments \\
10 & NC & Executive & Use policy & Use of High-Risk Applications Policy \\
11 & NC & Executive & Evaluation guidance & How Are Use Cases Evaluated \\
12 & NC & Executive & Press release & 2025: A Landmark Year for NCDIT in AI, Innovation, Connectivity and Security \\
14 & NC & Executive & Education guidance & NC Generative AI Implementation Recommendations for PK-13 Public Schools \\
15 & NC & Executive & Press release & NCDPI Partners With AI Innovation Index \\
16 & NC & Executive & Use policy & NCDHHS Acceptable Use Policy \\
17 & NC & Executive & Pilot report & Lessons from NC Treasurer's Office Generative AI Pilot with ChatGPT \\
49 & NC & Executive & Governance Roadmap & State of North Carolina AI Leadership Council: Statewide AI Strategic Roadmap \\

\addlinespace
\multicolumn{5}{@{}l}{\textbf{North Carolina: Legislative Branch}}\\
\addlinespace

18 & NC & Legislative & Bill & House Bill 375: AI/Ban Deceptive Ads \\
19 & NC & Legislative & Bill & House Bill 934: AI Regulatory Reform Act \\
20 & NC & Legislative & Law & Session Law 2025-38 House Bill 959 \\
21 & NC & Legislative & Law & Session Law 2024-37 House Bill 591 \\
22 & NC & Legislative & Bill & Senate Bill 640: AI Ethics and Literacy Across Education \\
23 & NC & Legislative & Bill & House Bill 1004: UNC AI \& Technology Hubs \\
24 & NC & Legislative & Bill & Senate Bill 746: Study Automation and the Workforce \\
25 & NC & Legislative & Bill & Senate Bill 735: AI Innovation Trust Fund \\
26 & NC & Legislative & Bill & Senate Bill 747: AI Learning Agenda \\
27 & NC & Legislative & Bill & House Bill 1036: AI Task Force/Funds \\

\addlinespace
\multicolumn{5}{@{}l}{\textbf{North Carolina: Judicial Branch}}\\
\addlinespace

28 & NC & Judicial & Ethics opinion & 2024 Formal Ethics Opinion 1: Use of AI in a Law Practice \\
30 & NC & Judicial & Judicial order & Administrative Order 25-09: AI in Superior Court Proceedings, Cabarrus County \\

\addlinespace
\multicolumn{5}{@{}l}{\textbf{North Carolina: Outside publishers (contextual sources)}}\\
\addlinespace

13 & NC & Outside & Contextual source & AI Spotlight: NC Improves Data-Driven Decision-Making with AI \\
29 & NC & Outside & Policy analysis & Artificial Intelligence, Local Orders, and the Courts \\

\addlinespace
\midrule
\multicolumn{5}{@{}l}{\textbf{Comparative State Corpus}}\\
\addlinespace

31 & AR & Executive & Report & Initial Report: Arkansas AI and Analytics Center of Excellence \\
32 & CA & Executive & Report & California Benefits and Risks of Generative AI Report \\
33 & CO & Legislative & Report & Report and Recommendations: AI Impact Task Force \\
34 & CT & Legislative & Report & Connecticut Artificial Intelligence Working Group Report \\
35 & GA & Legislative & Report & Final Report of the Senate Study Committee on AI, SR 476 \\
36 & IL & Legislative & Report & Report of the Generative AI and Natural Language Processing Task Force \\
37 & IN & Legislative & Report & Final Report: Artificial Intelligence Task Force \\
38 & KY & Legislative & Report & Findings and Recommendations of the AI Task Force \\
39 & MA & Executive & Report & Massachusetts AI Strategic Task Force 2024 Report to the Governor \\
40 & NJ & Executive & Report & Report to Governor on Artificial Intelligence \\
41 & NM & Legislative & Report & 2023 Interim Final Report \\
42 & OK & Executive & Report & Governor's Task Force on Emerging Technologies: AI Strategy for State Agencies \\
43 & OR & Legislative & Report & Final Report and Recommendations: Joint Task Force on AI \\
44 & SD & Legislative & Report & Study Committee on AI and Regulation of Internet Access by Minors \\
45 & TX & Legislative & Report & House Committee on AI and Emerging Technologies: Final Interim Report 2024 \\
46 & VT & Executive & Report & Artificial Intelligence Task Force: Final Report \\
47 & WA & Executive & Report & Inaugural Report of the Washington State AI Task Force \\
48 & WI & Executive & Report & Advisory Action Plan \\
\end{longtable}

\clearpage
\begin{longtable}{@{}L{0.8cm}L{3.7cm}L{2.2cm}L{4.3cm}L{2.5cm}@{}}
\caption{Provenance of the corpus documents. NCDIT is the NC Department of Information Technology and NCDPI is the NC Department of Public Instruction. Bill rows give the bill number, session, and edition read; edition 0 is the pre-filing draft. A dash means the document has no version or edition designation, and a living page is continuously updated, so its retrieval date identifies the text we read. The minutes for ID-4 are dated Nov 5, 2025, although the NCDIT file address reads Nov 4. H 1036 (ID-27) and H 591 (ID-21) are from the 2023--24 session; all other bills are from 2025--26. ID-14 is an NCDPI document hosted as a Google Doc. IDs 13 and 29 come from outside publishers.}
\label{tab:document-provenance}\\
\toprule
\textbf{ID} & \textbf{Issuing body} & \textbf{Issue date} & \textbf{Version or edition} & \textbf{Retrieved} \\
\midrule
\endfirsthead

\caption[]{Provenance of the corpus documents. Continued.}\\
\toprule
\textbf{ID} & \textbf{Issuing body} & \textbf{Issue date} & \textbf{Version or edition} & \textbf{Retrieved} \\
\midrule
\endhead

\midrule
\multicolumn{5}{r}{\emph{Continued on next page}}\\
\endfoot

\bottomrule
\endlastfoot

\multicolumn{5}{@{}l}{\textbf{North Carolina: Executive Branch}}\\
\addlinespace
1 & Office of the Governor & Sep 2, 2025 & --- & May 16, 2026 \\
2 & Office of the Governor & Sep 2, 2025 & --- & May 16, 2026 \\
3 & NCDIT & Undated & Living page & May 16, 2026 \\
4 & NCDIT & Nov 5, 2025 & --- & May 16, 2026 \\
5 & NCDIT & Jan 21, 2026 & --- & May 16, 2026 \\
6 & NCDIT & Aug 21, 2025 & v1.1 & May 16, 2026 \\
7 & NCDIT & Undated & Living page & May 16, 2026 \\
8 & NCDIT & Undated & Living page & May 16, 2026 \\
9 & NCDIT & Undated & Living page & May 16, 2026 \\
10 & NCDIT & Feb 6, 2025 & --- & May 16, 2026 \\
11 & NCDIT & Undated & Living page & May 16, 2026 \\
12 & NCDIT & Jan 12, 2026 & --- & May 16, 2026 \\
14 & NCDPI & Jan 16, 2024 & --- & May 16, 2026 \\
15 & NCDPI & Sep 25, 2025 & --- & May 16, 2026 \\
16 & NC Department of Health and Human Services & Sep 8, 2025 & v1 & May 16, 2026 \\
17 & NC Department of State Treasurer & Jul 1, 2025 & --- & May 16, 2026 \\
49 & NC AI Leadership Council, published by NCDIT & Jul 1, 2026 & As published; adopted Jun 9, 2026 & Jul 1, 2026 \\
\addlinespace
\multicolumn{5}{@{}l}{\textbf{North Carolina: Legislative Branch}}\\
\addlinespace
18 & NC General Assembly & Mar 13, 2025 & H 375, 2025--26; edition 1 & May 16, 2026 \\
19 & NC General Assembly & Apr 14, 2025 & H 934, 2025--26; edition 2 (bill filed Apr 10, 2025) & May 16, 2026 \\
20 & NC General Assembly & Jun 26, 2025 & H 959, 2025--26; S.L.\ 2025-38 & May 16, 2026 \\
21 & NC General Assembly & Jun 28, 2024 & H 591, 2023--24; edition 5; S.L.\ 2024-37 & May 16, 2026 \\
22 & NC General Assembly & Mar 25, 2025 & S 640, 2025--26; edition 0 (draft DRS35236-NG-83) & May 16, 2026 \\
23 & NC General Assembly & Apr 10, 2025 & H 1004, 2025--26; edition 0 (draft DRH10451-MTa-113A) & May 16, 2026 \\
24 & NC General Assembly & Mar 25, 2025 & S 746, 2025--26; edition 0 (draft DRS15268-LR-107A) & May 16, 2026 \\
25 & NC General Assembly & Mar 25, 2025 & S 735, 2025--26; edition 0 (draft DRS35250-MCa-139A) & May 16, 2026 \\
26 & NC General Assembly & Mar 25, 2025 & S 747, 2025--26; edition 0 (draft DRS45362-LR-142A) & May 16, 2026 \\
27 & NC General Assembly & May 2, 2024 & H 1036, 2023--24; edition 0 (draft DRH40574-LRa-133C) & May 16, 2026 \\
\addlinespace
\multicolumn{5}{@{}l}{\textbf{North Carolina: Judicial Branch}}\\
\addlinespace
28 & NC State Bar & Nov 1, 2024 & --- & May 16, 2026 \\
30 & Superior Court, Judicial District 25 (Cabarrus County) & Dec 8, 2025 & Revised order & May 16, 2026 \\
\addlinespace
\multicolumn{5}{@{}l}{\textbf{North Carolina: Outside publishers (contextual sources)}}\\
\addlinespace
13 & AI Center for Government (outside publisher) & Sep 10, 2025 & --- & May 16, 2026 \\
29 & UNC School of Government (outside publisher) & Apr 15, 2026 & --- & May 16, 2026 \\
\addlinespace
\multicolumn{5}{@{}l}{\textbf{Comparative State Corpus}}\\
\addlinespace
31 & Arkansas AI and Analytics Center of Excellence & Dec 6, 2024 & --- & May 16, 2026 \\
32 & California Government Operations Agency & Nov 1, 2023 & --- & May 16, 2026 \\
33 & Colorado Legislative Council, AI Impact Task Force & Feb 1, 2025 & --- & May 16, 2026 \\
34 & Connecticut AI Working Group & 2024 & --- & May 16, 2026 \\
35 & Georgia Senate Study Committee on AI & 2024 & --- & May 16, 2026 \\
36 & Illinois Generative AI and NLP Task Force & Dec 1, 2024 & --- & May 16, 2026 \\
37 & Indiana AI Task Force & Oct 10, 2024 & --- & May 16, 2026 \\
38 & Kentucky AI Task Force & Nov 13, 2024 & --- & May 16, 2026 \\
39 & Massachusetts AI Strategic Task Force & 2024 & --- & May 16, 2026 \\
40 & New Jersey AI Task Force & 2024 & --- & May 16, 2026 \\
41 & New Mexico Science, Technology \& Telecommunications Committee & 2023 & --- & May 16, 2026 \\
42 & Oklahoma Governor's Task Force on Emerging Technologies & Dec 29, 2023 & --- & May 16, 2026 \\
43 & Oregon Joint Task Force on AI & 2024 & --- & May 16, 2026 \\
44 & South Dakota Study Committee on AI and Regulation of Internet Access by Minors & 2024 & --- & May 16, 2026 \\
45 & Texas House Select Committee on AI and Emerging Technologies & Nov 1, 2024 & --- & May 16, 2026 \\
46 & Vermont AI Task Force & Jan 15, 2020 & --- & May 16, 2026 \\
47 & Washington State AI Task Force & Dec 1, 2024 & --- & May 16, 2026 \\
48 & Wisconsin Governor's Task Force on Workforce and AI & Jul 1, 2024 & --- & May 16, 2026 \\
\end{longtable}

\clearpage

\section{Model-Assisted Coding Workflow} 
\label{app:workflow}

Table~\ref{tab:workflow} summarizes the workflow. Stage 1 returned candidate topics for each whole document, and Stage 2 returned codebook matches, each with a context-specific explanation, a direct quotation as evidence, and a confidence score.

\begin{longtable}{@{}L{0.9cm}L{1.7cm}L{6.2cm}L{5.1cm}@{}}
\caption{Model-assisted coding workflow and validation.}
\label{tab:workflow}\\

\toprule
\textbf{Stage} & \textbf{Model} & \textbf{Model-assisted task and output} & \textbf{Human review} \\
\midrule
\endfirsthead

\caption[]{Model-assisted coding workflow and validation. Continued.}\\
\toprule
\textbf{Stage} & \textbf{Model} & \textbf{Model-assisted task and output} & \textbf{Human review} \\
\midrule
\endhead

\midrule
\multicolumn{4}{r}{\emph{Continued on next page}}\\
\endfoot

\bottomrule
\endlastfoot

1
& GPT-5.4
& Generated candidate inductive codes for each of the 49 documents from its full text. Each assignment included a high-level label, low-level label, definition, and confidence rating. This stage produced 8{,}589 candidate inductive assignments.
& Human review checked semantic support against the source document. \\

\addlinespace

2
& GPT-5.4
& Generated candidate deductive assignments to the five framework components, reading each whole document as a governance artifact.
& Human review checked that each quotation appeared in the document and supported the code. Retained assignments were linked to Niti Formalization, Nyaya Recourse, Pluralistic Governance of AI, Adaptive Governance of AI with Incremental Revision, and Bounded Rationality in AI Governance. \\

\addlinespace

3
& GPT-4o-mini
& Generated candidate semantic mappings between inductive labels and framework components, with primary and secondary mappings and confidence scores.
& Confidence scores were used only as triage aids. Human reviewers made final mappings. \\

\addlinespace

4
& Researcher review
& Reconciled mismatches and ambiguous cases, including criminal statutes and technical-object codes.
& Reviewers inspected flagged source passages and distinguished scope boundaries, mapping errors, and possible framework extensions. The final analytic dataset contains 8{,}589 mapped instances. \\

\end{longtable}

\newpage 

\section{Codebooks, Model Settings, and Crosswalk}
\label{app:codebooks}

\subsection{Theory-Derived Framework Codebook}

\begin{longtable}{@{}p{2.8cm}p{2.0cm}p{8.4cm}@{}}
\caption{Theory-derived framework codebook.}
\label{tab:codebook}\\

\toprule
\textbf{Framework component} & \textbf{Source theory} & \textbf{Operational definition} \\
\midrule
\endfirsthead

\caption[]{Theory-derived framework codebook. Continued.}\\
\toprule
\textbf{Framework component} & \textbf{Source theory} & \textbf{Operational definition} \\
\midrule
\endhead

\midrule
\multicolumn{3}{r}{\emph{Continued on next page}}\\
\endfoot

\bottomrule
\endlastfoot

Niti Formalization
& Sen's \textit{Niti}
& Formal rules, procedures, institutions, legal duties, oversight bodies, inventories, audits, transparency requirements, and accountability structures that make AI governance administrable. \\

\addlinespace

Nyaya Recourse
& Sen's \textit{Nyaya}
& Preconditions for public protection at three levels: recognition (harms, affected groups, benefit claims), tracking (outcomes), and recourse (complaint handling, recourse pathways, enforcement links, correction mechanisms). A passage at the first two levels offers no recourse. \\

\addlinespace

Pluralistic Governance of AI
& Stone's \textit{polis}
& Membership, power, influence, public-interest claims, market logic, information, co-operation, and representation in governance. The component asks who participates, whose evidence counts, and whose interests set the agenda. \\

\addlinespace

Adaptive Governance of AI with Incremental Revision
& Lindblom's muddling through
& Pilots, phased implementation, living documents, staged reform, partial revision, and pragmatic adjustment under uncertainty. \\

\addlinespace

Bounded Rationality in AI Governance
& Simon's bounded rationality
& Constraints of information, time, expertise, authority, staffing, and administrative bandwidth. The component includes satisficing, attention allocation, problem decomposition, local search, and routine reuse. \\

\end{longtable}

The dataset's component counts combine the three Nyaya levels, so claims about recourse in this paper rest on the provision-level reading in Tables~\ref{tab:matrix} and~\ref{tab:depth}, and no component count is read as evidence of recourse.  

\clearpage
\subsection{Pluralistic Governance of AI Subcodes}

\begin{longtable}{@{}p{3.3cm}p{10.6cm}@{}}
\caption{Pluralistic Governance of AI subcodes, grouped by the four pillars in Figure~\ref{fig:framework}.}
\label{tab:contested-standing-subcodes}\\

\toprule
\textbf{Subcode} & \textbf{Definition} \\
\midrule
\endfirsthead

\caption[]{Pluralistic Governance of AI subcodes, grouped by the four pillars in Figure~\ref{fig:framework}. Continued.}\\
\toprule
\textbf{Subcode} & \textbf{Definition} \\
\midrule
\endhead

\midrule
\multicolumn{2}{r}{\emph{Continued on next page}}\\
\endfoot

\bottomrule
\endlastfoot

\multicolumn{2}{@{}l}{\textit{Pillar a: stakeholders}} \\
\addlinespace

Membership
& Defining who belongs to the relevant community and who is excluded from participation, governance, benefits, or protections. \\

\addlinespace

Groups
& Recognizing groups and organizations as the primary actors through which political action and policymaking occur. \\

\addlinespace

Power
& Using influence, co-operation, loyalty, and collective action to subordinate individual self-interest to group or public interests. \\

\addlinespace

\multicolumn{2}{@{}l}{\textit{Pillar b: interests}} \\
\addlinespace

Motivation
& Accounting for both self-interest and altruism as motivations for political action and policymaking. \\

\addlinespace

Public interest
& Defining and contesting what serves the community, whether understood as shared individual interests, majority preferences, or collective well-being. \\

\addlinespace

Common interests
& Managing conflicts between private and collective interests when individual benefits create collective costs or collective benefits require private costs. \\

\addlinespace

Market logic
& Applying market-based reasoning, such as efficiency, cost-benefit analysis, or competitive advantage, to political decision-making. \\

\addlinespace

\multicolumn{2}{@{}l}{\textit{Pillar c: information}} \\
\addlinespace

Influence
& Changing what others believe, want, and do through persuasion, education, socialization, and strategic communication. \\

\addlinespace

Information
& Interpreting incomplete, ambiguous, or selectively presented information that depends on its source, framing, timing, and context. \\

\addlinespace

\multicolumn{2}{@{}l}{\textit{Pillar d: sustained co-operation}} \\
\addlinespace

Co-operation
& Building alliances and working collectively with others to pursue shared political goals and compete for influence. \\

\addlinespace

Loyalty
& Maintaining political relationships through trust, past obligations, and enduring commitments to allies. \\

\addlinespace

Community \& Collective Will
& Pursuing shared goals through collective action within a community. \\

\end{longtable}

\clearpage
\subsection{Bounded Rationality in AI Governance Subcodes}

\begin{longtable}{@{}p{3.0cm}p{10.0cm}@{}}
\caption{Bounded Rationality in AI Governance subcodes.}
\label{tab:capacity-subcodes}\\

\toprule
\textbf{Subcode} & \textbf{Definition} \\
\midrule
\endfirsthead

\caption[]{Bounded Rationality in AI Governance subcodes. Continued.}\\
\toprule
\textbf{Subcode} & \textbf{Definition} \\
\midrule
\endhead

\midrule
\multicolumn{2}{r}{\emph{Continued on next page}}\\
\endfoot

\bottomrule
\endlastfoot

Satisficing~\cite{simon1955behavioral, simon2000bounded}
& Accepting a workable option without searching for an optimal one. \\

\addlinespace

Attention allocation~\cite{jones2005politics, ocasio1997towards} 
& Prioritizing some problems over others because attention, expertise, and institutional bandwidth are limited. \\

\addlinespace

Problem decomposition~\cite{march1993organizations}
& Splitting a complex governance problem into smaller parts addressed by different actors or at different times. \\

\addlinespace

Local search~\cite{cyert1963behavioral}
& Looking to nearby examples, existing standards, peer states, templates, or available models for solutions. \\

\addlinespace

Routine reuse~\cite{nelson1985evolutionary}
& Reusing existing procedures, offices, definitions, or templates to reduce the burden of new governance work. \\

\end{longtable}

\clearpage 
\subsection{Interview-Derived Inductive Codes}

\begin{longtable}{@{}p{4.0cm}p{9.9cm}@{}}
\caption{Inductive codes from reflexive thematic analysis of the interview transcripts.}
\label{tab:inductive-codebook}\\

\toprule
\textbf{Code} & \textbf{Definition} \\
\midrule
\endfirsthead

\caption[]{Inductive codes from reflexive thematic analysis of the interview transcripts. Continued.}\\
\toprule
\textbf{Code} & \textbf{Definition} \\
\midrule
\endhead

\midrule
\multicolumn{2}{r}{\emph{Continued on next page}}\\
\endfoot

\bottomrule
\endlastfoot

Multilevel government authority
& Different levels of government hold different forms of enforcement authority and act on the same AI problems at once. \\

\addlinespace

Trust as precondition
& Participants treat trust and a sense of protection as conditions for policy uptake and describe the council as a trust-building institution under uncertainty. \\

\addlinespace

AI heterogeneity and policy granularity
& Grouping traditional machine learning, generative AI, and agentic AI under one framework creates tension between broad policy and field-specific guidance. \\

\addlinespace

Inevitability of AI
& Participants treat AI adoption as already underway and difficult to stop, regardless of state action. \\

\addlinespace

Experiential learning
& Participants treat hands-on use of AI tools as a way to build practitioner intuition and governance knowledge. \\

\addlinespace

Growth optimism
& Participants frame AI mainly through innovation, scaling, industry attraction, or economic development, and environmental or ethical concerns sometimes receive less attention. \\

\addlinespace

Anthropomorphization risk
& Participants are concerned that people may form emotional attachments to AI systems or treat them as persons, friends, or trusted companions. \\

\addlinespace

Reasoning by analogy
& Participants reason about AI governance through analogies to other technologies, institutions, or policy domains. \\

\addlinespace

Democratizing compute access
& Participants treat compute, tokens, or AI access as distributable public resources that could expand ordinary people's capabilities. \\

\addlinespace

AI reliability and verification
& Participants treat AI outputs as probabilistic and error-prone, and they hold that government action must be consistent, explainable, and correct. \\

\addlinespace

State as aggregator
& Participants position the state as a pooling or coordinating actor so that localities do not negotiate or govern AI alone. \\

\addlinespace

Symbolic policy
& Participants treat some measures as performative, misdirected, or weakly enforceable. \\

\addlinespace

External or non-traditional spaces as governance infrastructure
& Officials gather information and influence governance in spaces outside formal political settings. \\

\end{longtable}

\clearpage

\subsection{Model Settings and Prompts}
\label{app:prompts}

We made all API calls on July 7, 2026, with temperature 0.2 and the default top-p of 1.0 at every stage. API calls were performed on the full set of 49 documents as retrieved (Appendix~\ref{app:document-corpus}). No document was added, replaced, or updated after that date, and the human review and planning reading used the same retrieved texts. These calls coded document text only; binding status was recorded separately as of September 8, 2026 (Section 4). Prior to Stages 1 and 2, we utilized Azure AI Document Intelligence for Optical Character Recognition (OCR) as a preprocessing step to extract text from the raw documents, followed by validation of the extracted data. Stages 1 and 2 used \texttt{gpt-5.4} and sent the full text of one document in each call, with no chunking, no overlap, and no explicit cap on output tokens. No chunking of the documents was used because the document size ranged from 465 tokens (ID-22) to 61,967 tokens (ID-34), with a mean of about 9,705.61 which is less than one tenth of the input token size of the model used. Stage 3 used \texttt{gpt-4o-mini}, ran once per Stage 1 high-level and low-level codes paired with all Stage 2 high-level and low-level codes, and capped output at 700 tokens.

\paragraph{Inductive topic extraction.} System prompt: ``You return valid JSON only.'' User prompt:
 
\begin{promptbox}

\begin{verbatim}
You are a qualitative researcher analyzing policy documents.
Task: Extract the main themes and topics from this document.
For each identified topic:
1. Assign a HIGH-LEVEL CODE (broad theme)
2. Assign a LOW-LEVEL CODE (specific sub-topic)
3. Provide a DEFINITION (what this code means in context)
4. Rate CONFIDENCE (high/medium/low)
Document Text:
{document_text}
Output JSON format:
{
"topics": [
{
"high_level_code": "Theme name",
"low_level_code": "Specific topic",
"definition": "Clear explanation",
"confidence": "high"
}
]
}
\end{verbatim}
\end{promptbox}
 
\paragraph{Structured codebook matching.} System prompt: ``You are a precise JSON generator. Return ONLY valid JSON. Every match MUST have non-empty high\_level\_code, low\_level\_code, definition, evidence, and numeric confidence (0-1). Use ONLY codes from the provided codebook.'' User prompt:
 
\begin{promptbox}

\begin{verbatim}
You are a policy analyst applying a structured framework to documents.
CODEBOOK:
{codebook_text}
TASK:
Read this document and identify which codes from the codebook apply.
CRITICAL REQUIREMENTS:
1. ONLY use Theme and Code names EXACTLY as they appear in the codebook above
2. high_level_code and low_level_code fields MUST NOT be empty - always provide the
   exact Theme and Code names
3. For "definition": Write a CONTEXT-SPECIFIC explanation of HOW and WHY this code
   applies to THIS specific document. Do NOT just copy the codebook definition.
   Explain what you observed in the document that matches this code.
4. For "evidence": Extract direct quotes from the document that demonstrate this code
5. Provide confidence as a number between 0 and 1 (e.g., 0.9 for high confidence,
   0.5 for medium, 0.3 for low)
Document:
{document_text}
Output JSON (REQUIRED FORMAT):
{
"matches": [
{
"high_level_code": "Exact Theme name from codebook",
"low_level_code": "Exact Code name from codebook",
"definition": "Context-specific explanation of WHY/HOW this code applies to this
   document based on what you observed",
"evidence": "Direct quote from document showing this code applies",
"confidence": 0.9
}
]
}
If no codes apply, return: {"matches": []}
\end{verbatim}
\end{promptbox}
 
\paragraph{Crosswalk mapping.} System prompt: ``You return valid JSON only.'' User prompt:
 
\begin{promptbox}

\begin{verbatim}
You are an expert in qualitative coding and policy analysis.
TASK: Map a code from Framework A to the most similar code(s) in Framework B.
TARGET FRAMEWORK (Framework B):
{new_framework_ref}
CODE TO MAP FROM FRAMEWORK A:
High-Level: {old_high}
Low-Level: {old_low}
Definition: {old_def}
Document Frequency: {old_doc_count} documents
INSTRUCTIONS:
1. Find the PRIMARY best match in Framework B
2. Optionally identify SECONDARY match if the concept spans multiple codes
3. If no good match exists, return NO_MATCH
4. Provide confidence: high/medium/low
5. Explain your reasoning in 2-3 sentences
OUTPUT JSON:
{
"primary_high": "Target high-level code or NO_MATCH",
"primary_low": "Target low-level code or NO_MATCH",
"secondary_high": "Optional second match or null",
"secondary_low": "Optional second match or null",
"confidence": "high/medium/low",
"rationale": "Why these codes are conceptually similar"
}
\end{verbatim}
\end{promptbox}

\newpage 
\subsection{Crosswalk and Mismatch Resolution}
\label{app:inductive-mapping}

The crosswalk flagged 77 candidate assignments without a clean initial framework mapping. Reviewers resolved them by function (Table~\ref{tab:crosswalk-examples}). All 77 were assigned to a component, so mismatch resolution changed no counts: the final dataset retains all 8,589 initial inductive assignments.

\begin{longtable}{@{}L{2.8cm}L{4.1cm}L{3.2cm}L{3.8cm}@{}}
\caption{Resolution rules for initially unmatched inductive assignments.}
\label{tab:crosswalk-examples}\\

\toprule
\textbf{Passage function} & \textbf{Example content} & \textbf{Final component} & \textbf{Rationale} \\
\midrule
\endfirsthead

\caption[]{Resolution rules for initially unmatched inductive assignments. Continued.}\\
\toprule
\textbf{Passage function} & \textbf{Example content} & \textbf{Final component} & \textbf{Rationale} \\
\midrule
\endhead

\midrule
\multicolumn{4}{r}{\emph{Continued on next page}}\\
\endfoot

\bottomrule
\endlastfoot

Harm recognition
& Names sexual exploitation, CSAM, deepfakes, sextortion, intimate-image abuse, privacy loss, cybersecurity harm, unequal access, or workforce displacement.
& Nyaya Recourse
& The passage makes a public consequence visible, even if it does not specify measurement, recourse, or correction. \\

\addlinespace

Formal legal response
& Creates offenses, penalties, civil remedies, liability standards, court procedures, reporting duties, campaign rules, or whistleblower protections.
& Niti Formalization
& The passage establishes a rule, duty, remedy, procedure, or accountability structure. \\

\addlinespace

Object of governance
& Defines AI capabilities, legal categories, tool types, covered systems, synthetic media, or regulated uses.
& Niti Formalization
& The passage sets the scope of what the policy governs. \\

\addlinespace

Implementation constraint
& Describes enforcement difficulty, legal uncertainty, staffing limits, expertise gaps, agency burden, or interpretive uncertainty.
& Bounded Rationality in AI Governance
& The passage identifies what institutions can staff, interpret, enforce, or maintain. \\

\addlinespace

Technical boundary case
& Describes model architecture, compute, technical foundations, or infrastructure without connecting them to governance action.
& Retained because governance relevant
& Purely technical description was treated as outside the framework unless tied to authority, risk, deployment, oversight, or public consequence. \\

\end{longtable}
\clearpage

\subsection{What Each Analysis Contributed}
\label{sec:complement}

Table~\ref{tab:complement} lists what each analysis contributed. The model-assisted pass applied the same definitions to all 49 documents, and the 77 mismatches located passages the framework had to account for, including technical descriptions of models and infrastructure. A count of harm-recognition passages gives higher totals to documents that name many harms, so the counts alone could rank an advisory report above a bill that creates a remedy.

\begin{table*}[h]
\caption{Contributions of the model-assisted pass, the planning reading, and the interviews to each analytic question.}
\label{tab:complement}
\setlength{\tabcolsep}{4pt}
\renewcommand{\arraystretch}{1.12}
\begin{tabular}{p{4.6cm}p{4.6cm}p{4.6cm}}
\toprule
\textbf{Question} & \textbf{Model-assisted pass} & \textbf{Planning reading and interviews} \\
\midrule
What do the documents commit to? & Applied one codebook to all 49 documents & Separated obligations from intentions by reading operative language and instrument type \\
\addlinespace
Where does recourse appear? & Retrieved passages on harms and response mechanisms from all 49 documents & Recorded the date and binding status of each instrument (Section~\ref{sec:sequence}) \\
\addlinespace
Who influences decisions? & Retrieved passages naming publics across the corpus & Compared named publics with rosters, testimony lists, and interview accounts (Section~\ref{sec:standing}) \\
\addlinespace
Does the framework fit? & Generated inductive labels independent of the framework and flagged 77 mismatches & Added Bounded Rationality in AI Governance from interviews and resolved mismatches by function (Section~\ref{sec:vocabulary}) \\
\bottomrule
\end{tabular}
\end{table*}

\clearpage

\section{Framework Tables}
\label{app:framework-tables}
\small
\begin{longtable}{@{}L{0.15\linewidth}L{0.22\linewidth}L{0.15\linewidth}L{0.27\linewidth}L{0.15\linewidth}@{}}
\caption{How the framework follows from the requirements of studying AI
policy in formation. Each row gives a source theory, the condition AI
governance adds, and the resulting concept or indicator. Requirements 1
and 2 produce the central concepts, shown in bold, and requirements 3 to 6
produce operational indicators.}\label{tab:constructs}\\
\toprule
\textbf{Requirement} & \textbf{Source theory and the condition AI governance adds} & \textbf{Concept or indicator} & \textbf{Observable indicator} & \textbf{Component}\\
\midrule
\endfirsthead
\caption[]{Continued.}\\
\toprule
\textbf{Requirement} & \textbf{Source theory and the condition AI governance adds} & \textbf{Concept or indicator} & \textbf{Observable indicator} & \textbf{Component}\\
\midrule
\endhead
\midrule
\multicolumn{5}{r@{}}{\small\itshape Continued on next page}\\
\endfoot
\bottomrule
\endlastfoot
 
1.~Separating visibility to the state from recourse for the public &
Sen separates institutions from lives, and one AI instrument can inform the state without serving the public &
\textbf{Procedural visibility and public contestability} &
Whether the state, the public, or both can use what an instrument records; whether a provision is an agency duty or an affected-person right &
Niti Formalization; Nyaya Recourse \\ \addlinespace
 
2.~Evaluating policy across instruments and over time &
No single theory. Implementation research often follows one statute or program, and AI rules are split across instruments with different legal force &
\textbf{Binding sequence} &
Date written, binding status (in effect, enacted but not yet applicable, advisory, proposed, or absent), and who can enforce each duty &
All five \\*
 & & \textbf{Handoff specification} &
Whether a recommendation names who acts (implementation handoff) and the instrument that will enforce it (enforcement handoff) &
All five \\ \addlinespace
 
3.~Judging protection before outcomes exist &
Sen judges justice by realized lives, and officials write AI rules before anyone can observe them &
Preconditions of protection &
Three levels kept separate: recognition (harms, groups, benefit claims), tracking (outcomes), recourse (complaints, pathways, enforcement, correction) &
Nyaya Recourse \\*
 & & Depth of recourse specification &
Whether a harm is named, tracked, routed to an office, or tied to enforcement and correction &
Nyaya Recourse \\*
 & & Obligation rule &
Binding commitments separated from statements of intent &
Niti Formalization \\ \addlinespace
 
4.~Seeing publics who are absent from deliberation &
Stone centers on parties who organize and name an agent, and diffuse AI harms may have neither &
Standing gap &
Publics named in a document compared with its roster, witnesses, and drafters &
Pluralistic Governance of AI \\*
 & & Agent test &
Which roles the causal story identifies before a remedy is written: responsible institution, intentional perpetrator, regulated actor, affected party &
Pluralistic Governance of AI \\ \addlinespace
 
5.~Locating limits in institutions &
Bounded rationality describes limits on individuals and organizations, and budgets also set the capacity of AI councils and agencies &
Capacity signatures &
Satisficing, attention allocation, problem decomposition, local search, routine reuse &
Bounded Rationality in AI Governance \\ \addlinespace
 
6.~Knowing whether revision changes practice &
Lindblom assumes that decision-makers learn the consequences of each step, and AI harms are hard to observe &
Revision test &
Changes in institutional response compared with restated commitments &
Adaptive Governance of AI with Incremental Revision \\*
 & & Feedback condition &
Whether a route exists for public consequences to reach decision-makers &
Adaptive Governance of AI with Incremental Revision \\
\end{longtable}
\normalsize

\newpage 

\section{Additional Evidence}
\label{app:evidence}

\subsection{The Four Public Questions Across Instruments}

Table~\ref{tab:fourquestions} applies the four public questions to instruments from different policy trajectories to show how the test reads across them. These instruments follow different paths, so the table makes no claim about sequence.

\begin{table*}[h]
\caption{North Carolina instruments scored against the four public questions, with the edition we read and binding status as of September 8, 2026.}
\label{tab:fourquestions}
\setlength{\tabcolsep}{4pt}
\renewcommand{\arraystretch}{1.15}
\begin{tabular}{>{\raggedright\arraybackslash}p{2.6cm}>{\raggedright\arraybackslash}p{2.2cm}>{\raggedright\arraybackslash}p{2.0cm}>{\raggedright\arraybackslash}p{2.7cm}>{\raggedright\arraybackslash}p{2.5cm}>{\raggedright\arraybackslash}p{1.4cm}}
\toprule
\textbf{Instrument} & \textbf{Who is affected?} & \textbf{What harm is tracked?} & \textbf{Where can a person complain?} & \textbf{Who can correct it?} & \textbf{Status} \\
\midrule
Responsible Use Framework, version 1.1 (NC, ID-6) & Those who may be impacted & AI use & No complaint route specified & Agency; no right enforceable by a person & In effect as internal direction \\
\addlinespace
Executive Order No.~24 (NC, ID-1) & Implied by listed high-risk use cases & High-risk AI use & No individual complaint procedure specified & Agency AI oversight teams; no binding correction for individuals & In effect \\
\addlinespace
Roadmap (NC, ID-49) & Broad publics & Broad risks & Public feedback mechanism, read by staff & No binding correction; a person makes final benefits determinations; proposed AI Liaisons for accountability escalation & Advisory \\
\addlinespace
Senate Bill 735, edition 0 (NC, ID-25) & Public and users of covered models & AI safety incidents & No individual complaint procedure; developers report incidents to the attorney general & Attorney general enforcement & Proposed \\
\addlinespace
House Bill 934, edition 2 (NC, ID-19) & Persons aggrieved by deepfakes & Deepfake harms & Civil action & Courts & Proposed \\
\addlinespace
Session Law 2024-37, effective Dec 1, 2024 (NC, ID-21) & Identifiable minors; identifiable people in altered private images & AI-generated sexual imagery; altered private images & Criminal reporting; civil action & Courts and prosecutors; court may order the image destroyed & In effect \\
\bottomrule
\end{tabular}
\end{table*}

\clearpage

\subsection{Selected Roadmap Recommendations}

\textit{``Protect, prepare, and transform. Those are the three anchor points,''} a member said of the Roadmap's frame (P9). The Roadmap's letter reports 17 strategic goals, and its conclusion counts 25 recommendations (NC, ID-49). The Council identified secondary owners and funding sources, will meet quarterly, and \textit{``is ready to adjust and add to these recommendations over the next year''} (NC, ID-49). Table~\ref{tab:roadmap} lists selected recommendations, all of them advisory. The transparency goal also recommends that residents be able to opt out of data sharing beyond what program participation requires, and AI Liaisons would uphold whistleblower protections and serve as \textit{``the main point of contact for all accountability escalation''} (NC, ID-49). The Roadmap describes House Bill 959 as limiting student cellphone use and teaching social media literacy, and it names House Bill 375 and Senate Bill 738 as introduced bills (NC, ID-49).

\begin{table}[h]
\caption{Selected Roadmap recommendations, primary owners, and timelines (NC, ID-49).}
\label{tab:roadmap}
\setlength{\tabcolsep}{4pt}
\begin{tabular}{>{\raggedright\arraybackslash}p{6.6cm}>{\raggedright\arraybackslash}p{6cm}>{\raggedright\arraybackslash}p{1.7cm}}
\toprule
\textbf{Recommendation} & \textbf{Primary owners} & \textbf{Timeline} \\
\midrule
Minimum statewide trustworthy AI standards & Department of Information Technology & Sep 2026 \\
External reporting with public feedback & Department of Information Technology; Agency AI Oversight Committees & Dec 2026 \\
Benefits-program transparency and human determination & Department of Information Technology & Mar 2027 \\
Policy package on AI harms & Governor's Office; General Assembly & Mar 2027 \\
Policy package on worker displacement & Department of Commerce; General Assembly & Mar 2027 \\
Infrastructure-impact monitoring & Department of Environmental Quality; General Assembly; local governments & Jun 2027 \\
Cybersecurity threat monitoring and sharing & Department of Information Technology; Department of Public Safety; National Guard; local governments & Jun 2027 \\
Statewide AI contingency plan & Department of Information Technology; Department of Public Safety; National Guard & Sep 2027 \\
AI Liaisons in each Cabinet agency & Office of State Human Resources; agency leadership & Dec 2027 \\
Content verification standard & Department of Information Technology; agency communications teams & Mar 2028 \\
AI career transition program & Department of Commerce; NCWorks Career Centers; Community College System; local workforce boards & Dec 2028 \\
\bottomrule
\end{tabular}
\end{table}

\subsection{Remedies and Standing}

One member saw potential leaders \textit{``at every nook and cranny of all 100 counties in North Carolina'' }(P3). \textit{``It reminds me a lot of the broadband conversation we had 10 years ago,''} another said (P8). Participants expected workforce disruption and said that forecasting \textit{``where it's going to take place is a real piece of work that nobody really knows how to do right now''} (P5). California's warning concerns algorithmic error, a cause without intent, across six domains (CA, ID-32). A remedy needs someone to hold responsible and someone entitled to complain, and the causal story can supply both.

Staff read the input to the public feedback mechanism and share it with agency leadership. Members described review across subcommittees, with \textit{``a chance to review the others and comments''} (P6), and the Council worked through line-item edits on the Roadmap. Participants' accounts describe staff-led drafting and consensus that was less than complete, which limits the consensus building that Innes describes~\cite{innes1996planning}. Residents with limited prior exposure to algorithmic systems assessed and critiqued them quickly once researchers brought them into the discussion~\cite{eslami2025margins}, which suggests that the constraint is access to deliberation. Sen's criterion of public reasoning~\cite{sen2009idea, sen2008idea} applies to the standing gap: the reasoning in these documents came mainly from officials and industry experts. Public Citizen, an advocacy organization, reports that AI lobbying relationships rose 120 percent from 2022 to 2023, with 85 percent of the 2023 lobbyists representing corporate interests~\cite{cheng2024artificial}.

\end{document}